\documentclass[preprint,12pt]{elsarticle}

\usepackage{amssymb}

\usepackage{amsmath}
\usepackage{graphicx}
\usepackage{epstopdf}
\usepackage{url}
\usepackage{booktabs} 
\usepackage{xcolor} 
\usepackage{microtype} 
\usepackage{algorithm}
\usepackage{algorithmic}
\usepackage{makecell}
\usepackage{ulem}
\usepackage{comment}
\usepackage{multirow}
\usepackage{multicol}
\usepackage{ccaption}
\usepackage{hyperref}

\journal{Physica D: Nonlinear Phenomena}

\begin{document}

\begin{frontmatter}



\title{Symmetrical SyncMap for Imbalanced \\ General Chunking Problems}

\author[label1]{Heng Zhang}

\author[label1,label2]{Danilo Vasconcellos Vargas}

\affiliation[label1]{organization={Department of Information Science and Technology, Kyushu University},
            addressline={744 Motooka Nishi-ku}, 
            city={Fukuoka},
            postcode={819-0395}, 
            state={Fukuoka},
            country={Japan}}
            
\affiliation[label2]{organization={Department of Electrical Engineering and Information Systems, The University of Tokyo},
            addressline={Hongō 7-3-1}, 
            city={Bunkyo City},
            postcode={113-8654}, 
            state={Tokyo},
            country={Japan}}
\begin{abstract}
Recently, SyncMap pioneered an approach to learn complex structures from sequences as well as adapt to any changes in underlying structures. This is achieved by using only nonlinear dynamical equations inspired by neuron group behaviors, i.e., without loss functions.
Here we propose Symmetrical SyncMap that goes beyond the original work to show how to create dynamical equations and attractor-repeller points which are stable over the long run, even dealing with imbalanced continual general chunking problems (CGCPs). 
The main idea is to apply equal updates from negative and positive feedback loops by \textit{symmetrical activation}.
We then introduce the concept of \textit{memory window} to allow for more positive updates.
Our algorithm surpasses or ties other unsupervised state-of-the-art baselines in all 12 imbalanced CGCPs with various difficulties, including dynamically changing ones.
To verify its performance in real-world scenarios, we conduct experiments on several well-studied structure learning problems. The proposed method surpasses substantially other methods in 3 out of 4 scenarios, suggesting that symmetrical activation plays a critical role in uncovering topological structures and even hierarchies encoded in temporal data. 
\end{abstract}




\end{frontmatter}


\section{Introduction}
\begin{figure*}[ht]
    \centerline{\includegraphics[width = 0.99 \textwidth]{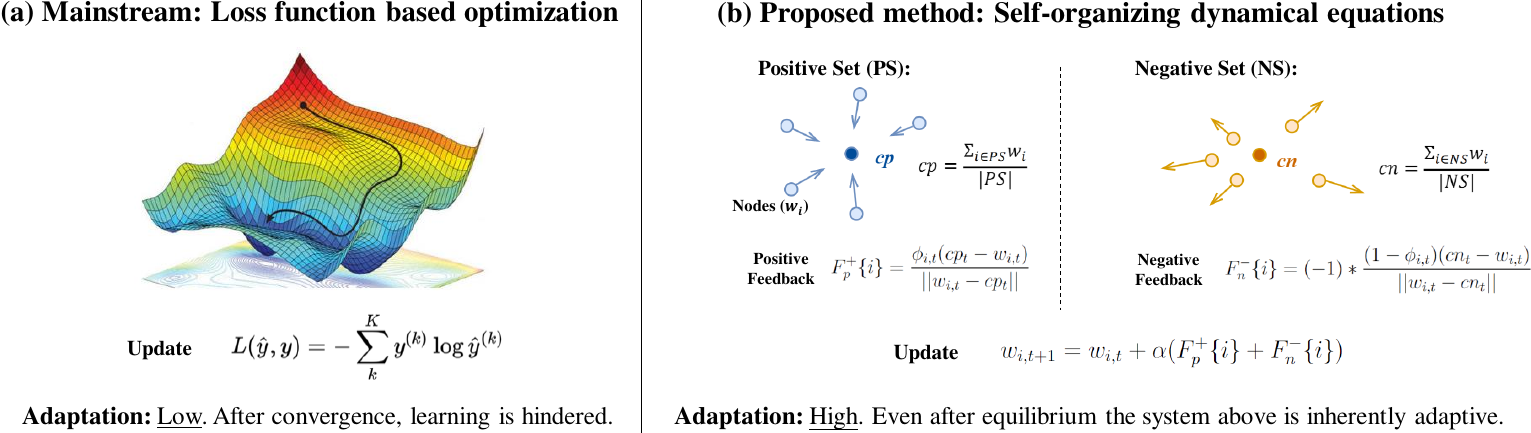}}
    \textcolor{black}{
    \caption{\textbf{Explanation of the proposed method compared with the mainstream.} 
    (a) Optimization based method. Loss function with stochastic gradient descent is mostly used to find a local optimum in the loss landscape (figure modified from \cite{amini2018spatial}). Once the minimum is reached, it can hardly adapt to the new environment.
    (b) Key concepts of the Symmetrical SyncMap under a new paradigm \cite{vargas2021syncmap}. It is a self-organizing dynamical system solely-based on nonlinear dynamical equations. After sufficient training, the system will enter an equilibrium where temporal structures lying in the input sequence are represented by the spacial correlation in the SyncMap state space. The system can quickly detect the structure changes and adapt to a new equilibrium.
    }
} 
    \label{fig:paradigm_compare}
\end{figure*}

Human brains have been proved to have unsupervised abilities to detect repetitive patterns in sequences involving texts, sounds and images \cite{orban2008bayesian,bulf2011visual,strauss2015disruption}. 
In the field of neuroscience, part of this behavior is known as chunking. Chunking has been verified in many experiments to play an important role in a diverse range of cognitive functions \cite{schapiro2013neural,yokoi2019neural,asabuki2020somatodendritic}.
Despite progress in neuroscience, chunking is still underexplored from computer science and machine learning.
Recently, Vargas et al. \cite{vargas2021syncmap} proposed the first learning of chunking based on nonlinear dynamical system and self-organization called \textit{SyncMap}. The authors also extended chunking problems into one called Continual General Chunking Problem (CGCP), which includes problems with diverse structures that can change dynamically throughout the experiments. 
\textcolor{black}{For the first time, SyncMap was shown not only able to uncover complex structures from sequential data, but also to adapt to continuously changing structures. 
Unlike the conventional machine learning methods that use loss and optimization functions to find the optimum, SyncMap try to reach an equilibrium rather than looking for optimum. This is achieved by nonlinear self-organizing dynamics that maps temporal input correlations to spacial correlations, where the dynamics are always updating with negative/positive feedback loops  (see Figure \ref{fig:paradigm_compare} for detail).
}
In this work, however, we identify problems in the original dynamics that lead to long-term instability,
and we further show that performances in imbalanced CGCPs are poor given the asymmetric number of updates, i.e., the number of negative updates is much bigger than that of the positive ones.

\subsection{Our contributions}
In this paper, we propose Symmetrical SyncMap, which can solve both of the problems above using symmetric selection of nodes and generalized memory window. Our main contributions are in the following:
\begin{itemize}

\item \textbf{More challenging problems with imbalanced structures.} We generalize chunking problems that are more difficult problems with imbalance data and structural input changes that require adaptation, called imbalanced CGCPs. While the original work fails in most of the problems due to the instability, Symmetrical SyncMap solves the problem at the foundation, keeping the final method concise and improving it in both accuracy and stability.
\item \textbf{Analysis and improvement of nonlinear equations.} We provide a deeper analysis of the original nonlinear equations. We identify the instability issue in the original work caused by the uneven updates from positive/negative feedback loops. We then propose \textit{symmetrical activation}, and further introduce the concept of \textit{memory window}, so that the system can have more updates from positive feedback loop while concurrently reducing the number of negative updates. 
\item \textbf{Achieving long-term stability and quick adaptation.} The symmetrical number of updates not only compensates when imbalanced chunks are presented, but also makes the algorithm stable over the long run and reaches an equilibrium quickly in changing environments. By showing that equilibrium and self-organization can appear only with dynamical equations and without optimization/loss functions, the biggest motivation from this paper is realizing how the substantial improvements make the new learning paradigm very adaptive and precise. 
\item \textbf{Superior performance in imbalanced CGCPs and real-world scenarios.} Our algorithm surpasses or ties other state-of-the-art baselines in all 12 imbalanced CGCPs with various difficulties, including dynamically changing ones. The stability of maintaining equilibrium, the preciseness, as well as the adaptability to changes in underlying structures, also highlight its potential in real-world applications, as verified by the results of two well-studied benchmarks.
\end{itemize}


\section{Chunking}
Natural neural systems are well known for the unsupervised adaptivity, since they can self-organize by many mechanisms for several purposes on many timescales \cite{lukovsevicius2012reservoir}. One of the mechanisms is chunking, which can be described as a biological process where the brain attains compact representation of sequences \cite{estes2007can, ramkumar2016chunking}. 
Specifically, long and complex sequences are first segmented into short and simple ones, while frequently repeated segments are concatenated into single units \cite{asabuki2020somatodendritic}. 
This can be seen as a complexity reduction for temporal information processing and associated cost \cite{ramkumar2016chunking}.
Chunking is an extremely multidisciplinary problem.
Albeit our focus is more on neuroscience and machine learning perspectives, earlier algorithms proposed for solving chunking problems are from linguistics and include PARSER \cite{perruchet1998parser}. It performs well in detecting simple chunks, but fails when the probability of state transition are uniform \cite{schapiro2013neural}. 
In recent years, an unsupervised learning method for chunking was proposed by using dual reservoir computing, where two reservoirs supervise each other to mimic the other's outputs to the same temporal input \cite{asabuki2018interactive}, yet the model requires the number of chunks to be known previously. 
A neuro-inspired sequence learning model, Minimization of Regularized Information Loss (MRIL) was proposed by applying a family of competitive network of two compartment neuron models that aims to predict its own output in a type of self-supervised neuron \cite{asabuki2020somatodendritic}. 
Albeit the interesting paradigm, MRIL has been shown unstable even for problems in which it performs reasonably well. Very recently, a self-organizing learning paradigm (called SyncMap) has been proposed, which surpassed MRIL in all scenarios \cite{vargas2021syncmap}.

\section{Continual General Chunking Problems}
Continual General Chunking Problem (CGCP) was recently first proposed by \cite{vargas2021syncmap}.
The author generalized various problems from neuroscience to computer science, including chunking, causal and temporal communities and unsupervised feature learning of time sequences, into one single problem.
CGCP is considered as extracting co-occurring states from time sequences, in which the generation process (i.e., data structure) can change over time. 
\textcolor{black}{
The input generation process is defined by a random walk on a graph, transitioning by a first-order Markov chain. 
In the settings, a state can be associated to either a fixed chunk or a probabilistic chunk.
The details are explained as follows:}

\textcolor{black}{
\textbf{State space.} Consider a graph $\mathbf{G}$ that consists of $N$ possible nodes (i.e., state variables). The input time sequence is then generated by performing a random walk on the graph, in which the set of possible states in the time sequences can be considered as the state space, and the transition from one state to another is governed by the transition probabilities.
}

\textcolor{black}{
\textbf{Transition probabilities.} The probability of moving from state $i$ to state $j$ in one step is defined by the transition matrix $\mathbf{P}$, where $\mathbf{P}[i][j]$ is the probability of transitioning from state $i$ to state $j$. 
For a random walk on a graph, this probability is $1/d[i]$ if there is an edge between $i$ and $j$ and $0$ otherwise. 
Here, $d[i]$ is the degree of node $i$, which is the number of edges connected to node $i$. This can be written as:
}

\textcolor{black}{
\begin{equation}
    \mathbf{P}[i][j] = \mathbf{A}[i][j] / d[i],
\end{equation}
}

\noindent where
\textcolor{black}{
$i, j = 1, 2, ..., N$.
Note that this is the matrix form of the transition rule defined earlier, where $\mathbf{A}$ is the adjacency matrix of the graph $\mathbf{G}$, and $d[i] = \sum \mathbf{A}[i][j]$, for $j = 1,..., N$. 
Simple put, our assumption is the equal state transition probability that if the node $i$ is connected to $k$ other nodes, the walker has a $1/k$ probability of going to each of them.
}

\begin{figure}[ht]
    \centering
    \includegraphics[width=0.99\linewidth]{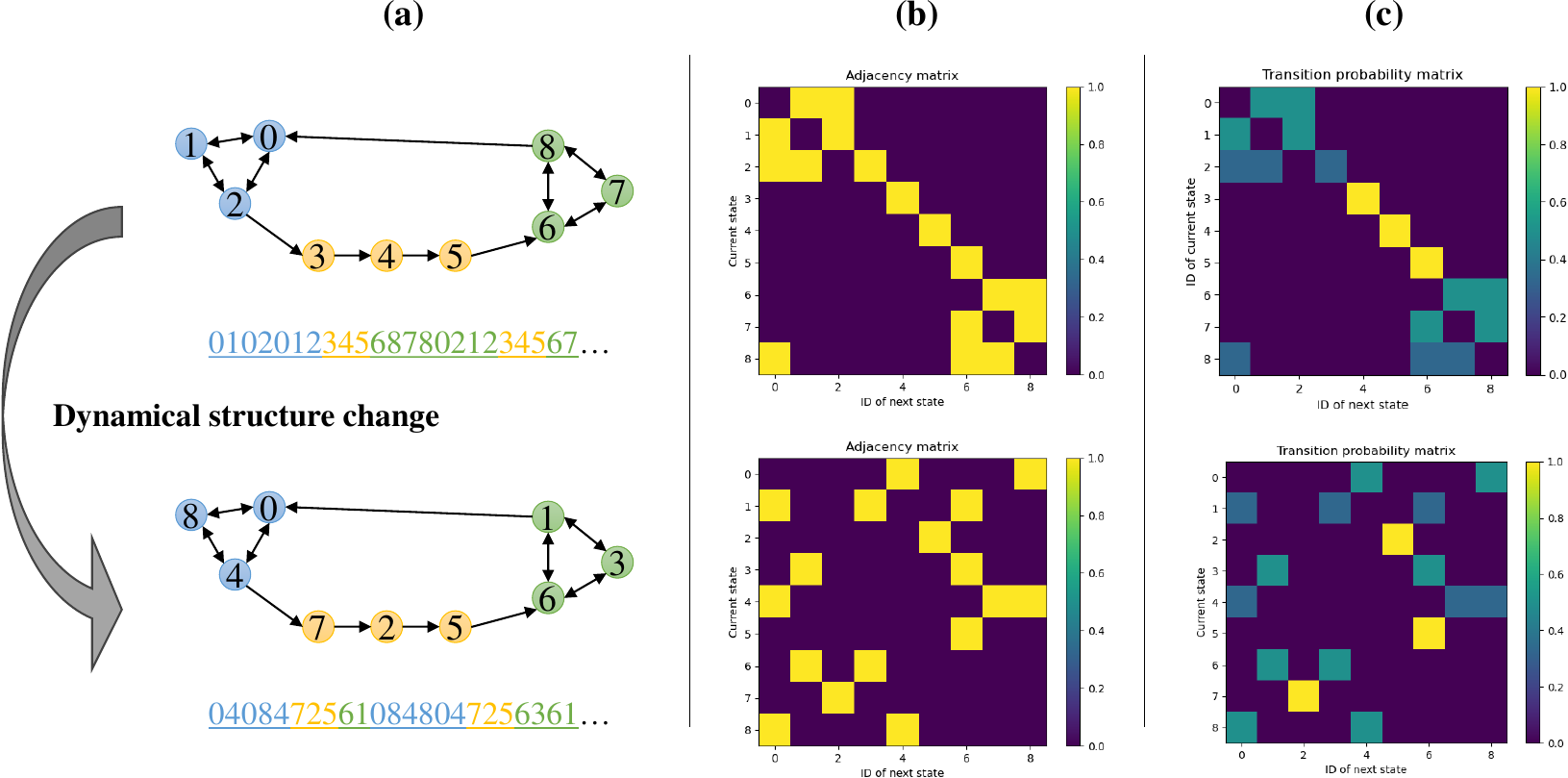}
    \caption{
    \textcolor{black}{
    \textbf{Data generation process of CGCP problem.} (a) The data structure of a toy example used in \cite{vargas2021syncmap}. This is a mixed CGCP problem with 2 probabilistic chunks (012 and 678) and a fixed chunk (345). The series of number below is an example of the generated sequence that is further fed into SyncMap. (b) The corresponding adjacency matrix of (a). To perform a random walk on the graph, (b) is then converted into (c) transition probability matrix.
    Moreover, CGCP considers the dynamical environment change, as the data structure used to generate input can change over time, which requires the algorithm to have a good adaptation ability. Here, the structure is changed to having 2 probabilistic chunks (048 and 136) and a fixed chunk (725).
    }
    }
    \label{fig:cgcp_overview}
\end{figure}

\textcolor{black}{
\textbf{Fixed chunk} problem (see Figure \ref{fig:cgcp_overview}) refers to the situation that the next state $i+1$, with respect to the current state $i$, is deterministic within a chunk. For example, if $a$ and $b$ are two continuous elements of a fixed chunk with direction $a$ to $b$, then we have: 
\begin{equation}
    \mathbf{P}[a][b]=1 \quad | \quad \mathbf{P}[a][j]=0, j\neq b.
\end{equation}
}

\textcolor{black}{
\textbf{Probabilistic chunk.} 
More realistically, the probabilistic chunk (see Figure \ref{fig:cgcp_overview}) can be seen as \textit{community} in a graph. In CGCP, the generation process of probabilistic chunks is typically defined as a subset of nodes that are more densely connected to each other than to nodes in other communities in the graph. 
In other words, nodes within the same community have more edges between them, indicating a higher \textit{internal} degree of interaction or similarity, while \textit{external} degree that cross communities are relatively rare.
Formally, let $C$ be a probabilistic chunk, with $i$ being a state in $C$, $j$ being a state not in $C$, and $\mathbf{A}$ remaining as the adjacency matrix of the graph. The internal degree $d_{in}(C)$ of $C$ is given by:
}

\textcolor{black}{
\begin{equation}
    d_{in}(C) = 0.5 * \sum_i \sum_j \mathbf{A}[i][j] \quad | \quad \{i, j\} \in C,
\end{equation}
\noindent where the factor of 0.5 is there because each internal edge is counted twice in the sum.
The external degree $d_{ex}(C)$ is given by:
}

\textcolor{black}{
\begin{equation}
    d_{ex}(C) = \sum_i \sum_j \mathbf{A}[i][j] \quad | \quad i \in C, j \notin C.
\end{equation}
\noindent For any probabilistic chunk, we have:
\begin{equation}
    d_{in}(C) > d_{ex}(C).
\end{equation}
}

\textcolor{black}{
A toy example is shown in Figure \ref{fig:cgcp_overview}. It describes the input generation process of (1) initializing a graph, (2) obtaining adjacency matrix, (3) obtaining transition probability matrix, and (4) input sequence generated by performing a random walk on the graph based on the transition probability matrix. 
}

\textcolor{black}{
Beyond generalizing chunking problems to fixed and probabilistic chunks, CGCP also considers their continual variations. This is motivated by the constant adaptation observed by neural cells that can relatively switch behavior quickly in different environments \cite{dahmen2010adaptation}. In this case, the data structure can change over time, conferring a harder albeit realistic setting (see Figure \ref{fig:cgcp_overview}).
}

\section{SyncMap}
\begin{figure}[H]
    \centering
    \includegraphics[width=0.75\linewidth]{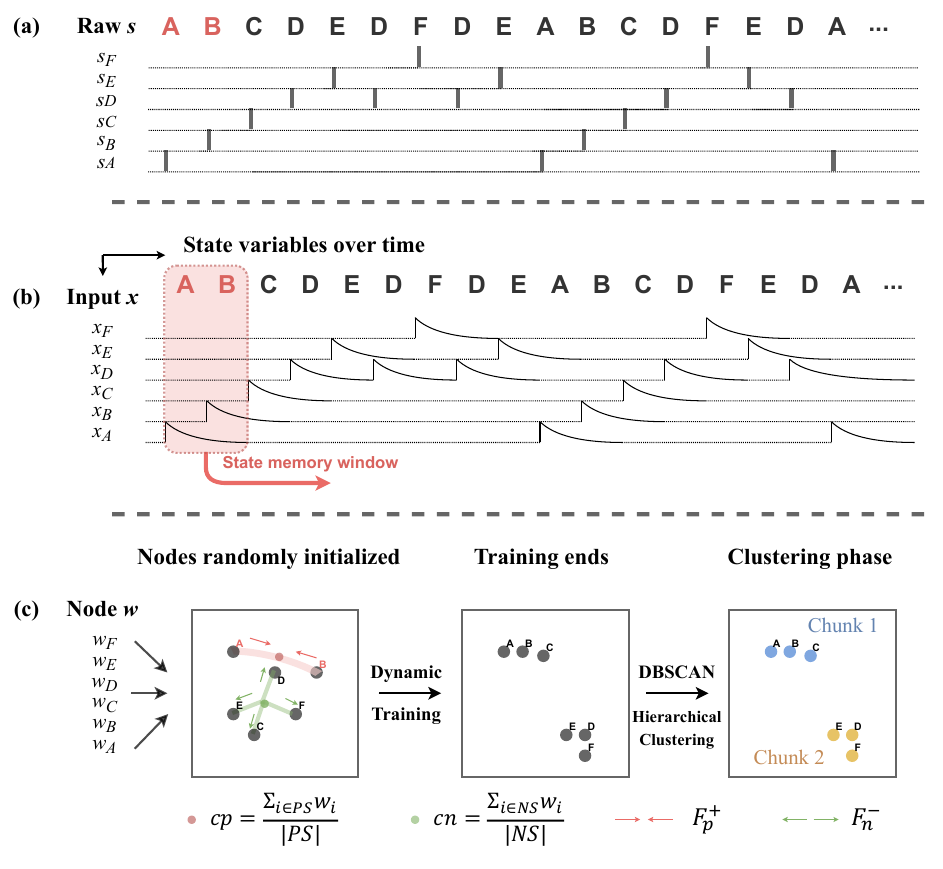}
    \caption{
    \textbf{SyncMap overview.}
    The chunking procedures of SyncMap. 
    (a) Raw input sequence $S$. Gray columns represent a state variable is activated.
    (b) Sequences with state variables are first encoded as exponentially decaying input $\boldsymbol{x}_t$. Here, the state memory $m$=2.
    (c) SyncMap state space. Weight nodes $w_{i}$ in SyncMap's dynamic are randomly initialized (left). The dynamic is then trained by Equations \ref{eq4}-\ref{eq7}. Finally, in the clustering phase, DBSCAN/Hierarchical clustering is applied to obtain chunks/communities (right). }
    \label{fig:syncmap_overview}
\end{figure}

SyncMap is a self-organizing nonlinear dynamical system proposed by \cite{vargas2021syncmap}. It solves CGCP by creating nonlinear mapping that encodes the temporal correlation (chunks) as spatial distance between nodes. 
In SyncMap's dynamic (i.e., a generated map space), nodes which are activated together tend to be grouped as chunks, while nodes that do not activate together will be pulled away from each other. The algorithm is explained in detail in the following, with a brief example shown in Figure \ref{fig:syncmap_overview}.

\textbf{Input encoding.} 
\textcolor{black}{
The raw data sequence of a CGCP problem is initially generated as the temporal sequences shown in Figure \ref{fig:cgcp_overview}(a).
However, before feeding the input into SyncMap, an encoding process is needed.
Consider a matrix-form of the raw input sequence of state variables $\boldsymbol{S}=[\boldsymbol{s}_1,\boldsymbol{s}_2,...,\boldsymbol{s}_t,...,\boldsymbol{s}_{\tau}]$, where $\tau$ is the sequence length.
$\boldsymbol{s}_t={\{s_{1,t},...,s_{N,t}\}}^{T}$ is a vector at time step $t$, and its elements $s_{i,t}$ ($i=1,...,N$) hold constrain $s_{i,t}\in \{0,1\}:\sum^{N}_{i=1}s_{i,t}=1$, where $N$ is the total number of possible states. 
This is the formal mathematical form of the sequence shown in Figure \ref{fig:cgcp_overview}. Since the data generation process undergoes the first-order Markov chain, at each time step, only one element in $\boldsymbol{s}_t$ equals to 1 (current state) while the others are 0, similarly to the one-hot encoding in some classification tasks. 
Then, the input is encoded as an exponentially decaying vector $\boldsymbol{x}_t={\{x_{1,t},...,x_{N,t}\}}^{T}$ having the same shape as $\boldsymbol{s}_t$:
}
\begin{equation} \label{eq1}
    x_{i,t}=\begin{cases}s_{i,t_a}*e^{-0.1*(t-t_a)}, & t-t_a<m*tstep \\
    0, & otherwise
    \end{cases} ,
\end{equation}

\noindent where $t_a$ is the last state transition to state $\boldsymbol{s_i}$, and $m$ is the state memory. 
Specifically, state transitions happen every $tstep$ step (time delay), and variables that have their activation period greater than $m*tstep$ are set to 0. $m$ and $tstep$ are set at 2 and 10 in the original work. See Figure \ref{fig:syncmap_overview} the encoded exponentially decaying input, where the sequence \textit{abcdedfdeabcfeda} has a fixed chunk \textit{abc} and a probabilistic chunk \textit{edf}.

\textbf{Training dynamic.} 
\textcolor{black}{SyncMap can be seen as a nonlinear dynamical system. The dynamic resides in the training phase, where it maps the encoded sequences into a dynamically changing spatial space (the Euclidean space). This mapping is nonlinear, as the temporal correlations in the sequences are captured as spatial correlations in the SyncMap space by using the following update equations.
To illustrate, we generate weight nodes $w_{i,t}$ in SyncMap's map space to obtain pair tuple ($x_{i,t}$, $w_{i,t}$). Nodes are first randomly initialized in a $k$ dimensional Euclidean space. Note that weight node $w_{i,t}\in \mathbb{R}^k$ is a point in SyncMap's map space, and it can also be considered as a vector. 
}

In every iteration when a new input vector $\boldsymbol{x}_{t}$ comes in, all its elements $x_{i,t}$, together with the corresponding nodes $w_{i,t}$, are divided into two sets according to the threshold value $a$: (1) positive set $PS_t=\{i|x_{i,t}>a\}$ including activated or recently activated states, as well as (2) negative set $NS_t=\{i|x_{i,t}\leq a\}$ including non-recently activated states. 
The original SyncMap used $a$ directly at $0.1$. In this paper we introduce threshold value $a$ which allows us to achieve more general state memory implementation.

Inside the space, the centroids of $PS_t$ and $NS_t$ sets are calculated as follows if and only if the cardinality of both sets are greater than one in this iteration (i.e., \(|PS_t|>1\) and \(|NS_t|>1\)):
\begin{equation} \label{eq4}
    cp_t=\frac{\sum_{i\in PS_t}w_{i,t}}{|PS_t|}, \quad
\end{equation}
\begin{equation} \label{eq5}
    cn_t=\frac{\sum_{i\in NS_t}w_{i,t}}{|NS_t|},
\end{equation}
where $cp_t$ and $cn_t$ are the centroids of $PS_t$ and $NS_t$ respectively. 
Finally, node $w_{i,t}$ corresponding to each input $x_{i,t}$ is updated as follow:
\begin{equation} \label{eq6}
    \phi_{i,t} =
    \begin{cases} 
     {1, \ i\in PS_t}\\{0, \ i\in NS_t} 
    \end{cases}, 
    \alpha=\begin{cases}
    {\alpha, \ i\in PS_t\cup NS_t}\\{0, \ otherwise}
    \end{cases},
\end{equation}

positive feedback $F^{+}_{p}\{i\}$:
\begin{equation}
\label{eq:pfb}
    F^{+}_{p}\{i\} = \frac{\phi_{i,t}(cp_t-w_{i,t})}{||w_{i,t}-cp_t||},
\end{equation}

negative feedback $F^{-}_{n}\{i\}$: 
\begin{equation}
\label{eq:nfb}
    F^{-}_{n}\{i\} = (-1) * \frac{(1-\phi_{i,t})(cn_t-w_{i,t})}{||w_{i,t}-cn_t||},
\end{equation}


\begin{equation} \label{eq7}
    w_{i,t+1}=w_{i,t}+\alpha(F^{+}_{p}\{i\} + F^{-}_{n}\{i\}),
\end{equation}
\noindent where $\alpha$ is the learning rate and $||$·$||$ is the Euclidean distance. 
Subsequently, updated nodes are normalized to be within a hyper-sphere having radius of 10 at the end of the iteration.

\textbf{Clustering phase.} 
\textcolor{black}{
SyncMap forms a $k$-dimensional map during dynamic training, which has the number of nodes equal to the total number of possible states $N$.
After training, the dynamic map is fixed and a clustering process is performed to read out the chunks/communities detected by SyncMap.
In this work, DBSCAN \cite{schubert2017dbscan} is used in the clustering phase.
DBSCAN is a widely-used clustering algorithm that operates on a Euclidean space and does not require predefining the number of clusters. 
It identifies dense regions based on a user-defined density parameter $eps$ and minimum cluster size $mc$, revealing patterns and relationships within the data. This approach enhances model interpretability by extracting clusters of varying sizes and shapes from the fixed map.
Alternatively, one can apply other clustering algorithms such as hierarchical clustering \cite{murtagh2012algorithms} to obtain a hierarchical structure of the detected chunks. We used hierarchical clustering in the two real-world community detection benchmarks in Section \ref{sec:real world}.
}

\begin{figure}[ht]
    \centering
    \includegraphics[width=0.7\linewidth]{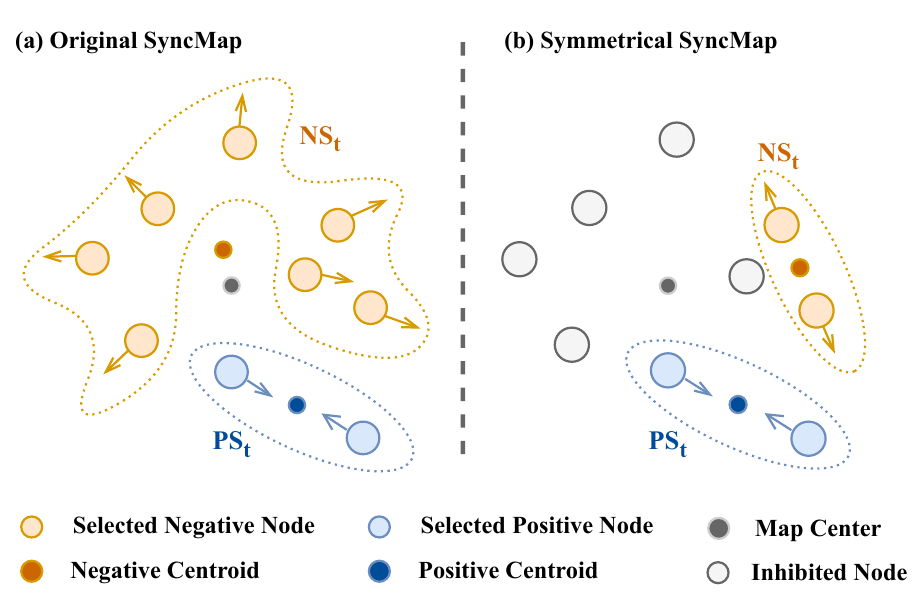}
    \caption{
    \textbf{Comparison of node's update.}
    (a) Illustration of the instability in SyncMap. 
    In the original work, the dynamical equations are strongly influenced by negative nodes, since the cardinality of all non-activating nodes $NS_t$ are usually much greater than that of activating nodes $PS_t$ (e.g., 7 : 2). 
    (b) The proposed symmetrical activation. By applying stochastic selection, equal number of positive and negative nodes are activated in each iteration (e.g., 2 : 2). }
    \label{fig:comparison_stoSelect}
\end{figure}

\textbf{Limitations of SyncMap.}
Although SyncMap shows capabilities to address all kinds of CGCP, one crucial issue is the instability of its dynamic in the long term. 
This is due to the asymmetric number of updates with respect to positive and negative nodes.
Figure \ref{fig:comparison_stoSelect}(b) shows how this happens with an example of nine nodes in 2-D SyncMap. 
The fixed state memory ($m$ = 2) results in an uneven update of positive (2) and negative (7) nodes, i.e., the dynamic's update is more influenced by negative feedback loop, which causes an undesirable convergence in the long run.

\begin{figure*} 
    \centering
    \includegraphics[width=0.99\linewidth]{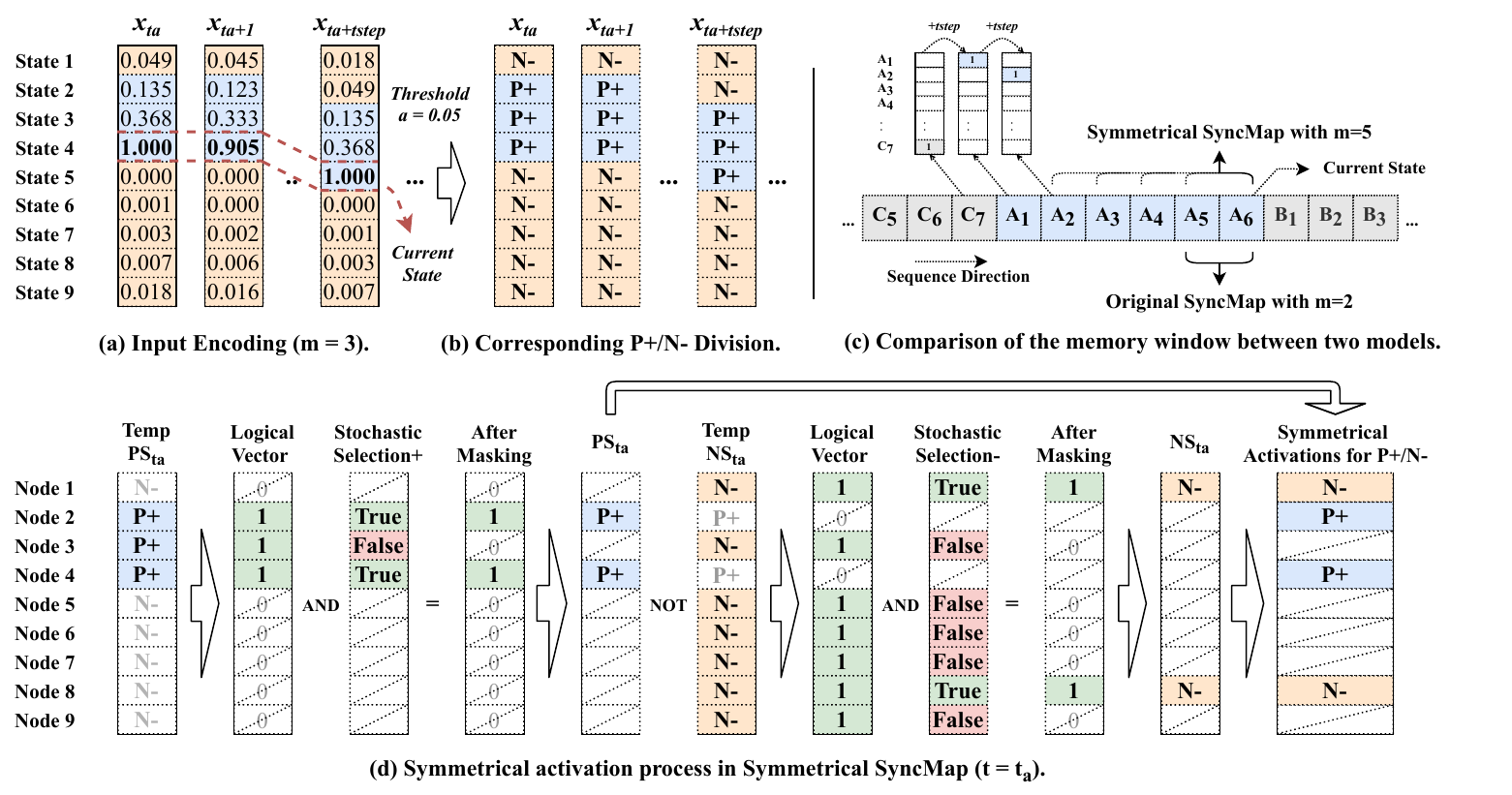}
    \caption{
    \textbf{General workflow of symmetrical activation.}
    (a) Exponentially decaying input sequence.
    (b) Sequence after the process of positive and negative nodes' division (represented as logical vectors).
    (c)  Comparison of the memory window between the original and Symmetrical SyncMap. State $A_i$, $B_i$ and $C_i$ belong  to  three  fixed chunks. The original SyncMap deterministically activates $m$ most recent states (i.e., $A_5$ and $A_6$), while Symmetrical SyncMap generalizes the state memory to have a larger memory window (e.g., from $A_2$ to $A_6$ when $m$=5) for stochastically selecting and activating positive nodes.
    (d) Process of stochastic selection to achieve symmetrical activation. First, we randomly lose sight of some nodes in $PS_t$ set (i.e., instead of activating all positive nodes, we stochastically select some of them to activate). This is achieved by an AND operation to the input and a masking vector having random logical values at each time step.
    Next, for stochastic selection of the negative nodes in $NS_t$ set, we use a masking vector similarly to that in positive part, and end up with activating equal number of positive and negative nodes in every iteration.
    }
    \label{fig:PDF_methodology_all}
\end{figure*}

\section{Symmetrical SyncMap}
Inspired by how \textit{neural efficiency} influences brain activation by focusing the energy on smaller brain areas \cite{neubauer2009intelligence}; here we propose an algorithm called Symmetrical SyncMap to better solve CGCP, particularly the imbalanced chunking problems. 
The main idea is to use symmetrical positive and negative activations. 
In other words, we try to reduce the number of activated negative nodes while selecting and activating more positive nodes in every iteration, thus balancing the updating rates in negative and positive feedback loops. 
To achieve symmetrical activation, we introduce \textit{stochastic selection} and \textit{memory window}.

\subsection{Memory Window: Generalizing the State Memory}
We introduce memory window to our algorithm by generalizing the state memory $m$, which allows a wider window for updates from the positive feedback loop, thus helping to capture the bigger chunks.
This is achieved by tuning the threshold value $a$ as mentioned in SyncMap's definition, i.e., any nodes $w_{i,t}$ having its corresponding input value $x_{i,t}$ greater than $a$ will be divided into $PS_t$ set and vice versa (a true or false logical operation). 
With a pre-defined $tstep$, one can easily adjust threshold $a$ to control the state memory $m$ ($tstep$=10, $a$=0.05 and $m$=3 in Figure \ref{fig:PDF_methodology_all}(a)). 
\textcolor{black}{A detailed analysis of how the thresholding works is shown in \ref{Appendix:Parameter Sensitivity Anaylsis}.}

\subsection{Symmetrical Activation}
Symmetrical activation is the core of our proposed algorithm, where equal number of positive and negative nodes are selected to activate at each time step. We propose stochastic selection to select nodes without bias in $PS_t$ and $NS_t$ sets. Details are shown in Figure \ref{fig:PDF_methodology_all}. 

\textbf{Stochastically select nodes into $\boldsymbol{PS_t}$ set.}
With a pre-defined state memory $m$, we first obtain the temporary $PS_{temp}$ set in a same way the original SyncMap obtains $PS_t$ (i.e., $PS_{temp}$ includes $m$ positive nodes, and $PS_{temp} \subseteq W_t$, where $W_t=\{w_{i,t}|i=1,...,n\}$ is the set including all nodes).
Then, we apply stochastic selection to select positive nodes into $PS_t$ (i.e., a sampling process). 
Whether to enable stochastic selection at this particular time step is determined by a probability parameter $Pr\in [0,1]$. 
In other word, if stochastic selection were enabled, we randomly select 2 positive nodes and ``inhibit" (ignore) other $m-2$ nodes in $PS_{temp}$, with the probability of $Pr$ when state memory $m>2$; otherwise we select all $m$ nodes.
When $m=2$, stochastic selection is not used and two most-recent states are selected.
Afterwards, $PS_t$ is updated, which only includes those activated positive nodes ($PS_{t} \subseteq PS_{temp}$). 
Additionally, we give an analysis of $Pr$ in  \ref{Appendix:Parameter Sensitivity Anaylsis}.

\textbf{Stochastically select nodes into $\boldsymbol{NS_t}$ set.}
After obtaining the above $PS_t$, we define the temporary negative set $NS_{temp}$=$W_t$\ -\ $PS_t$.
One may notice that there is a chance that some nodes in $PS_{temp}$ could potentially be sampled as nodes in $NS_{temp}$. This is desirable as it introduces a more uniform selection process and produces more robust results.
Next, we again use the stochastic selection for sampling several negative nodes in $NS_{temp}$ set. The number of negative nodes being selected is symmetrically equal to the cardinality of $PS_t$ (i.e., $|NS_t|=|PS_t|$). 
After this step, the $NS_t$ set is updated ($NS_{t} \subseteq NS_{temp}$, 
see the right part of Figure \ref{fig:PDF_methodology_all}(d) for example). 

The remaining steps follow the Equations \ref{eq4}-\ref{eq7}. We calculate a moving average of 10000 steps of nodes' position and use it for DBSCAN, instead of applying DBSCAN to the map at a ``snapshot'' time step in the original work.
Algorithmic description is shown in  \ref{Appendix:Algorithmic}.

\section{Related Works}


\textbf{Time-series Clustering.} Time series data is defined as a sequence of continuous, real-valued elements, usually with high dimensionality and large data size \cite{aghabozorgi2015time}. As a subroutine in unsupervised sequence processing, time-series clustering aims to uncover patterns, usually in very large sequential datasets that cannot be manually handled. 
This can be found in some articles applying competition-based self-organizing maps (SOMs) \cite{kohonen1990self} and their variations \cite{vannucci2018self,fortuin2018som}, which are well-suited for clustering time series but not capable of chunking time series. 
In other words, these SOMs were not designed to find the underlying structures of sequences and correlation between variables, therefore, their objectives are different. 

\textbf{Word Embeddings.} In the field of natural language processing, word embedding algorithms generally transforms texts and paragraphs into vector representations \cite{khattak2019survey, bojanowski2017enriching, peters2018deep}.
FastText enriched the word vector with subword information \cite{bojanowski2017enriching}, whereas ELMo \cite{peters2018deep} and BERT \cite{devlin2018bert} aimed to represent word by contextualized word embeddings.
Chunking problems presented here are related to some of them, such as a prediction-based Word2vec embedding algorithm \cite{mikolov2013efficient} that transforms texts into a vector space representation and can be combined with clustering to deal with chunking problems. 
Therefore, Word2vec is used in the experiments.  

\textbf{Representation Learning and Communities Detection.}
The problem of finding probabilistic chunks refers to a random walk over a graph with several chunk structures; in which the possibility of transition to an internal state within a chunk is higher than that of transition to an external state belonging to other chunks.
Such graph structures mentioned above can be seen as communities \cite{radicchi2004defining}, which are most-studied by recent representation learning algorithms such as DeepWalk \cite{perozzi2014deepwalk} and Graph Neural Networks (GNNs) \cite{kipf2016semi}. 
More related, the Modularity Maximization \cite{newman2004finding, tang2009relational} uses eigen-decomposition performed on the modularity matrix to learn vertex representation of community.
By using the adjacency matrix (transition probability matrix) to convert sequential data to graph structure, Modularity Maximization can also deal with chunking problems via random walk over the generated graphs. 
Although there exists newer modularity-based algorithms which try to optimize the pioneering work, such as Louvain method \cite{blondel2008fast}, their objective is mostly to reduce the computational cost.
Therefore, we use the original Modularity Maximization in the comparison.

\section{Experiments of CGCP problems}
We evaluate the proposed Symmetrical SyncMap with 13 tests, including (i) a long-term behavior analysis, and (ii) 12 imbalanced CGCP with fixed, probabilistic and mixed chunks, as well as their continual variations.
Among the large number of clustering quality measurements, we used Normalized Mutual Information (NMI) \cite{studholme1999overlap} for measurement (see  \ref{Appendix:nmi} for math detail). 
NMI ranges between 0 and 1, and the higher the score, the better the chunking performance. 
We did 30 trials for each experiment. Results of NMI are shown in Figure \ref{fig:balanced_fixed} and Table \ref{table2}. We used a t-test with p-value of 0.05 to verify if the best result is statistically significantly different from the others (statistical results are in  \ref{Appendix:Statistical Tests}). 
An ablation study and computational time analysis are also investigated in \ref{Appendix:Ablation Study} and \ref{Appendix:Computational Time Anaylsis}. 
\textcolor{black}{
Codes of the experiment can be found at \url{https://github.com/Roger2148/Symmetrical_SyncMap}.
}

\subsection{Long Term Behavior Analysis Experiment}
To evaluate the behavior of the algorithm over the long run, we set up this experiment.
The problem considered here is a continual changing environment with ten fixed chunks, each containing six different states. Transitions between chunks happen at the end of a chunk sequence (i.e., after the sixth state variable presented inside a chunk). A chunk can transit to any other chunk with equal probability.
Sequence length was set to $\tau$=600000. At time step $t$=0 the first environment was initialized. After $t$=300000 the problem changed to the second environment by re-assigning all 60 states into new ten chunks. 
We applied the original SyncMap and the proposed one in this experiment. 
We used the same parameter settings of the two models, where $\alpha$=0.001*$n$, $k$=2, $eps$=1 and $mc$=2. Besides, $m$=3 and $Pr$=30\% were used for Symmetrical SyncMap.

\begin{figure*}[ht]
    \centering
    \includegraphics[width=0.99\linewidth]{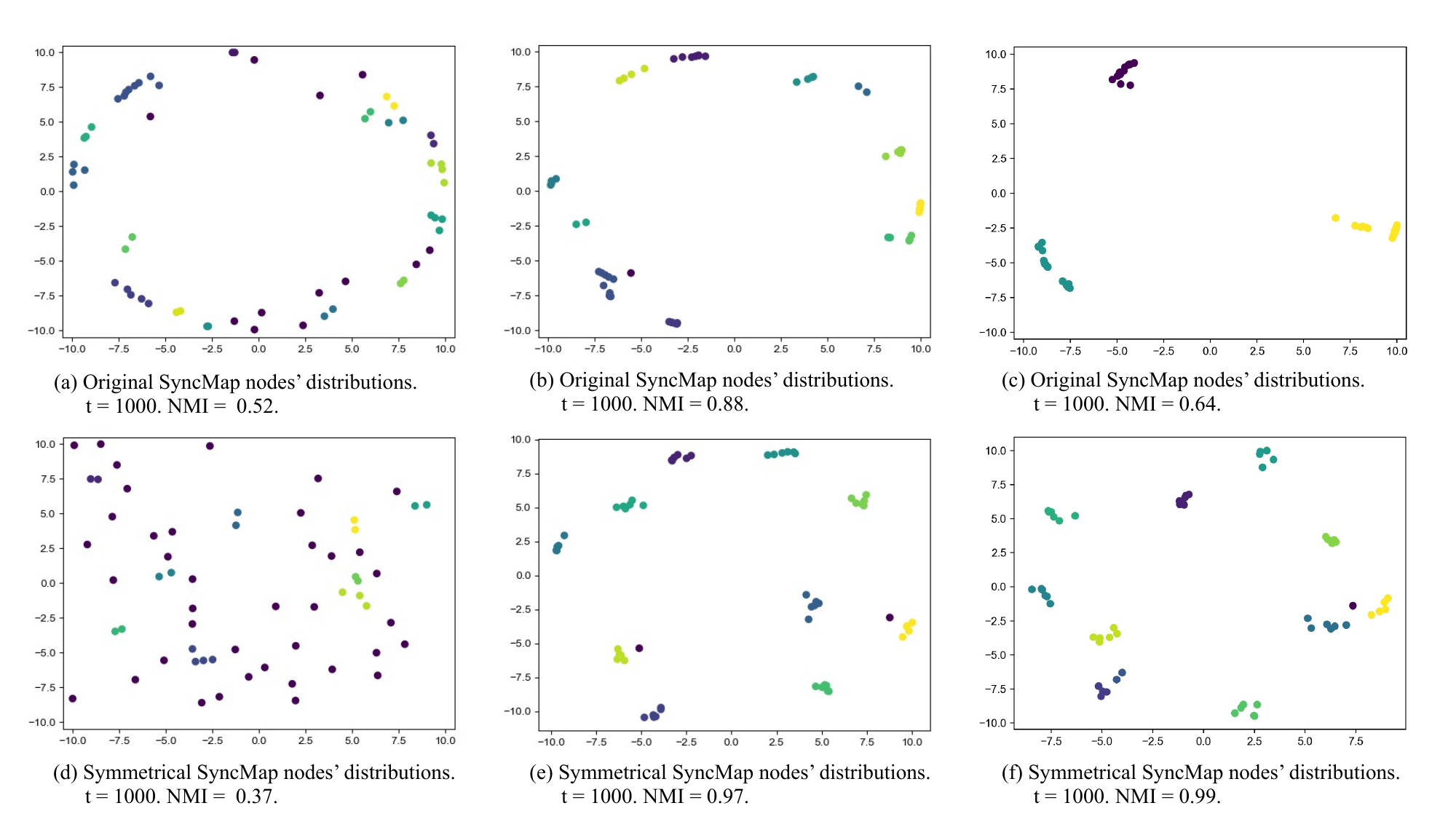}
    \caption{
    \textbf{Long term behavior analysis of the original SyncMap and Symmetrical SyncMap.} (a) NMI over time of two models. Data are mean with s.t.d (error bar). Statistics are in  \ref{Appendix:Statistical Tests}. 
    (b) For comparison, we show a single trial performed by two models, initialized identically. (c) Nodes' distribution at different time steps. Colors of the nodes indicate the true labels of chunks.
    }
 \label{fig:balanced_fixed}
\end{figure*}

Results in Figure \ref{fig:balanced_fixed} show that Symmetrical SyncMap reaches near the optimal performance in this experiment. 
By applying symmetrical activation, Symmetrical SyncMap can have long-term stability while keeping NMI near 1.0. 
In contrast, the NMI of the original SyncMap reaches the peak near 0.88 at $t$=70000 and decreases constantly afterwards. 
After environment changes, Symmetrical SyncMap detects new chunks and reaches to a new equilibrium quickly, while the original SyncMap performs poorer and becomes unstable in the long run.

\subsection{Imbalanced CGCP Problems}
\label{sec:experiment}
\textbf{Baselines and Parameter Settings.}
We test several imbalanced CGCP problems and their continual variations by using \textit{Symmetrical SyncMap, SyncMap, Modularity Maximization (Modularity Max), Word2vec and MRIL}. 
In detail, Symmetrical SyncMap's parameters were set to $\alpha$=0.001*$n$, $k$=3, $m$=3, $Pr$=30\%, $eps$=4.5 and $mc$=2. We conducted a parameter sensitivity analysis shown in  \ref{Appendix:Parameter Sensitivity Anaylsis}.
For the original SyncMap, we used $k$=3, $m$=2,  $eps$=4.5 and $mc$=2. 
Regarding the Modularity Max, we first converted the input sequence to transition probability (TP) matrix, and then used the TP matrix to generate a graph for communities detection. 
To evaluate how a word embedding algorithm would fair in CGCP, a skip-gram Word2vec algorithm was modified to suit in the context of CGCP. 
Here, a latent dimension of 3 and an output layer with softmax were used, and the output size is equal to the inputs. 
Learning rate was set at 0.001 and batch size was 64 with a mean squared error as loss. A window of 100 steps (equivalent to 10 state transitions) was used to compute the output probability of skip-gram. 
Regarding the MRIL, we used 5 output neurons for all experiments, with the learning rate of 0.001. 
We gathered the output neurons showing correlation larger than 0.5, detecting chunks by assigning an index of groups that maximally respond to each input. 
The input sequences of all baselines were the same exponential decaying input as used in Symmetrical SyncMap. 

\textbf{Problem Settings (\ref{Appendix:exmaple CGCP}).}
We first consider several environments which consist of 3 different sizes of chunks: big, moderate and small chunks. 
Specifically, the big chunk has 20 state variables, while the moderate and small chunks have 10 and 5 respectively. 
Based on the chunk settings, we then designed three types of imbalanced problems:
(i) Two big and one small chunks (20-20-5).
(ii) One big, one moderate and one small chunks (20-10-5).
(iii) One big and two small chunks (20-5-5). 
We tested these three types of imbalanced problems with three different structure settings: fixed, probabilistic and mixed chunks tests.
\textcolor{black}{The adjacency and transition probability matrices of two example structures are shown in Figure \ref{fig:imbalanced cgcp mats}. 
Details can be found in  Figure \ref{fig:chunk_problems_all} in  \ref{Appendix:exmaple CGCP} for the examples of the complete structures.
}
Regarding the mixed tests, two probabilistic chunks and one fixed chunk were presented in each environment, where the order of chunks in the input sequence was specified as: $1^{st}$ probabilistic to fixed to $2^{nd}$ probabilistic chunk. 
Sequence length $\tau$ is set at 200000 for all types of test.

\begin{figure*}[ht]
    \centering
    \includegraphics[width=0.99\linewidth]{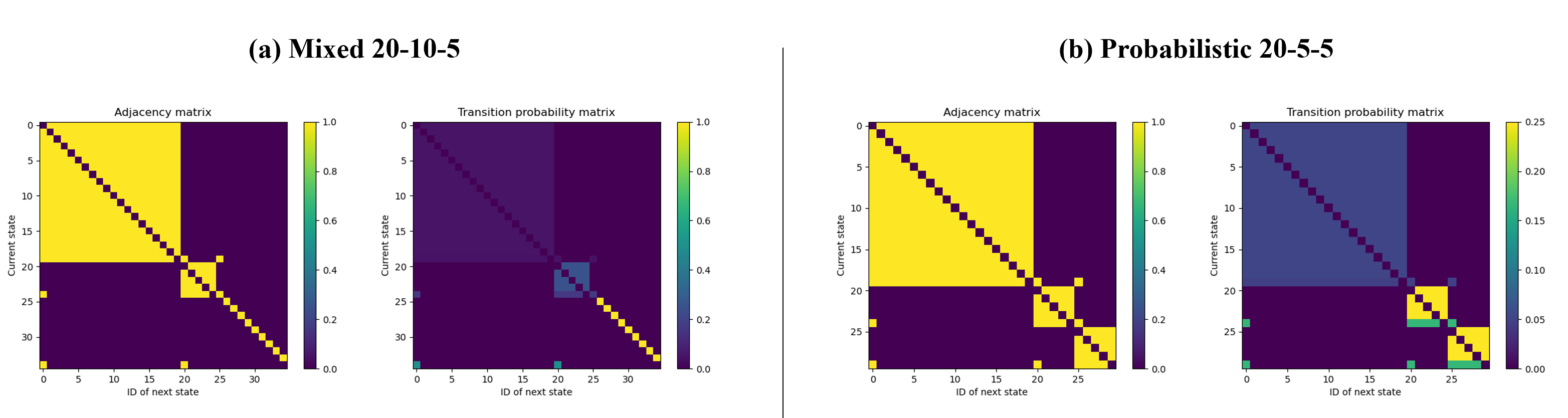}
    \caption{
    \textcolor{black}{
    \textbf{Examples of the imbalanced structures used in the experiments.} 
    (a) Imbalanced mixed 20-10-5 problem, where the fixed chunk has 10 states shown at the bottom-right on both adjacency matrix (left) and transition probability matrix (right).
    (b) Imbalanced 20-5-5 problem.
    }
    }
 \label{fig:imbalanced cgcp mats}
\end{figure*}

\textbf{Dynamical Continual Variation.}
Three dynamical variations of the above-mentioned problems were presented: continual fixed, continual probabilistic and continual mixed. Sequence length was set to $2\tau$. States were permuted between chunks: at time step $t$=0 the first type of problem was 15-15-5 (see the problem formalism in previous subsection), after $t$=$\tau$ the second type of problem was 20-10-5.

\begin{table*}[ht]
    \centering
    \resizebox{0.99\textwidth}{!}{%
    \begin{tabular}{c |c c c |c c c | c c}
    \hline
    \multirow{2}{*}{\textbf{Algorithm}} & \multicolumn{3}{c|}{\textbf{Fixed}} & \multicolumn{3}{c|}{\textbf{Probabilistic}} &
    \multicolumn{1}{c}{\textbf{SBM Network}}\\
    & 20-20-5&20-10-5&20-5-5&20-20-5&20-10-5&20-5-5 & \multicolumn{1}{c}{25-30-35}\\
    \hline
      Modularity Max &0.67±0.0&	0.73±0.03&	0.64±0.02&	0.96±0.04	&\textbf{1.0±0.0}&	\textbf{1.0±0.0} & \multicolumn{1}{c}{\textbf{0.99±0.02}}\\
      Word2vec &0.49±0.05&	0.57±0.07&	0.56±0.06&	0.70±0.04&	0.77±0.09&	0.62±0.08 & \multicolumn{1}{c}{0.84±0.03}\\
      MRIL& 0.25±0.09&	0.38±0.12&	0.36±0.11&	0.43±0.14&	0.39±0.07&	0.24±0.04 & \multicolumn{1}{c}{0.46±0.10}\\
      Original SyncMap  & 0.93±0.12&	0.75±0.08&	0.63±0.11&	\textbf{1.0±0.0}&	0.81±0.04&	0.64±0.08 & \multicolumn{1}{c}{\textbf{1.0±0.0}}\\
      \textbf{Ours}: Symmetrical SyncMap & \textbf{1.0±0.0}&	\textbf{1.0±0.0}&	\textbf{0.93±0.08}	&\textbf{1.0±0.0}	&\textbf{1.0±0.0}	&\textbf{1.0±0.0} & \multicolumn{1}{c}{\textbf{1.0±0.0}}\\
      
      \hline
    \multirow{2}{*}{\textbf{Algorithm}} & \multicolumn{3}{c|}{\textbf{Mixed}} & \multicolumn{3}{c|}{\textbf{Continual 15-15-5 to 20-10-5}}  & \multicolumn{1}{c}{\textbf{Whales' Song}} \\
    &20-20-5&20-10-5&20-5-5&Fixed&Prob.&Mixed & 8-8-6 \\
          \hline
      Modularity Max &0.69±0.05&	0.78±0.05 &	0.89±0.06&	0.69±0.02 &	0.70±0.05&	0.64±0.02 & \textbf{0.43±0.04} \\
      Word2vec &0.66±0.07&	0.60±0.06&	0.73±0.05&	0.45±0.04&	0.60±0.04&	0.65±0.04 & 0.19±0.05\\
      MRIL& 0.20±0.05&	0.20±0.05&	0.16±0.03&	0.38±0.13&	0.59±0.02&	0.55±0.04 & 0.39±0.08 \\
      Original SyncMap  & \textbf{0.84±0.08}&	0.83±0.0&	0.64±0.07&	0.72±0.07&	0.82±0.04&	0.83±0.0 & 0.33±0.14\\
      \textbf{Ours}: Symmetrical SyncMap & \textbf{0.87±0.09}&	\textbf{0.90±0.06} &	\textbf{0.95±0.04}&	\textbf{1.0±0.0}&	\textbf{1.0±0.0} & \textbf{0.95±0.06} & 0.35±0.05\\
      \hline
    \end{tabular}}
    \caption{
    \textbf{NMI results.} A comparison is shown over Modularity Max, Word2vec, MRIL, original SyncMap and Symmetrical SyncMap in imbalanced and real-world CGCPs. The best and the non-statistically different results are in bold. Data are mean±s.t.d. Details of the statistical t-tests (p values) are presented in  \ref{Appendix:Statistical Tests}.
    }
    \label{table2}
\end{table*}

\textbf{Results Overview.} 
The proposed algorithm Symmetrical SyncMap learns nearly the optimal solutions in all imbalanced CGCPs. It surpasses or ties other algorithms in all tests (Tables \ref{table2}).
Modularity Max performs the second best, in which it wins or ties the others in 2 out of 3 probabilistic CGCP tests.
Word2vec achieves relatively higher NMI in probabilistic CGCPs than other problem structures, whereas MRIL performs the worst overall in all tests. 
The original SyncMap performs good in 20-20-5 CGCPs, yet performance decrease is witnessed as more chunks become smaller. 


\textbf{Symmetrical SyncMap}, with its inherent adaptivity, performs significantly better than all other competitive algorithms, particularly in continual variations (i.e., dynamical CGCPs where environment can change). 
The proposed wider memory window and symmetrical activation allow capturing states in big chunks compactly, while at the same time the stochastic selection with suitable $Pr$ helps to separate small chunks (See the learned maps and $Pr$ analysis in \ref{Appendix:learned map} and \ref{Appendix:Parameter Sensitivity Anaylsis}), thus keeping the balance between dealing with small and big chunks.
In all probabilistic CGCP tests, it produces very distinct chunks and learns the best solution (i.e., NMI=1.0). 
The performance downgrades slightly during the more challenging mixed CGCP tests, due to an extra imbalanced frequency issue: the fixed chunk inserted between two probabilistic chunks has lower frequency to appear in the sequences.
Having said that, the proposed algorithm still outperforms the others in all mixed CGCPs with a very big lead.

\textbf{Original SyncMap} performs relatively better in 20-20-5 type CGCPs, with the steady decrease in 20-10-5 and 20-5-5 ones. In fixed CGCPs, it makes distinct clusters for smaller chunks, yet fails to group nodes of the big chunk together. 
In probabilistic and mixed tests, 
nodes belong to smaller chunks are merged into one cluster in almost every individual trial (see Figures in  \ref{Appendix:learned map}).

\textbf{Modularity Max} shares the highest NMI score with Symmetrical SyncMap in two probabilistic CGCP tests. However, this graph-based algorithm does not perform well in other imbalanced CGCPs.
A possible reason is that fixed chunks are less likely to appear in usual problems faced by Modularity Max, thus leading to a problem bias.
It is worth noting that comparing the results of Modularity Max to the other algorithms in dynamical CGCPs is not fair, since a TP matrix would record all the occurrence of state variables; thus, passing a continual changing TP matrix is not inherently suitable for Modularity Max, leading to worse results in continual structures than that in static graphs.

\textbf{Word2vec} creates maps in which nodes are more dispersed than that produced by SyncMap, thus making clustering difficult.
It performs better in probabilistic chunk tests than fixed and mixed ones (see Figures \ref{fig:UB_figures} in  \ref{Appendix:learned map}). 
For the continual problems, Word2vec lacks the ability of adaptation, thus showing the overall lower NMI scores.
\textbf{MRIL}
fails to detect imbalanced chunks with large number of state variables, and therefore it performs the worst in all tests. 
Increasing the number of output neurons may improve the performance of fixed chunk tests.


\begin{figure*}[ht]
    \centering
    \includegraphics[width=0.7\linewidth]{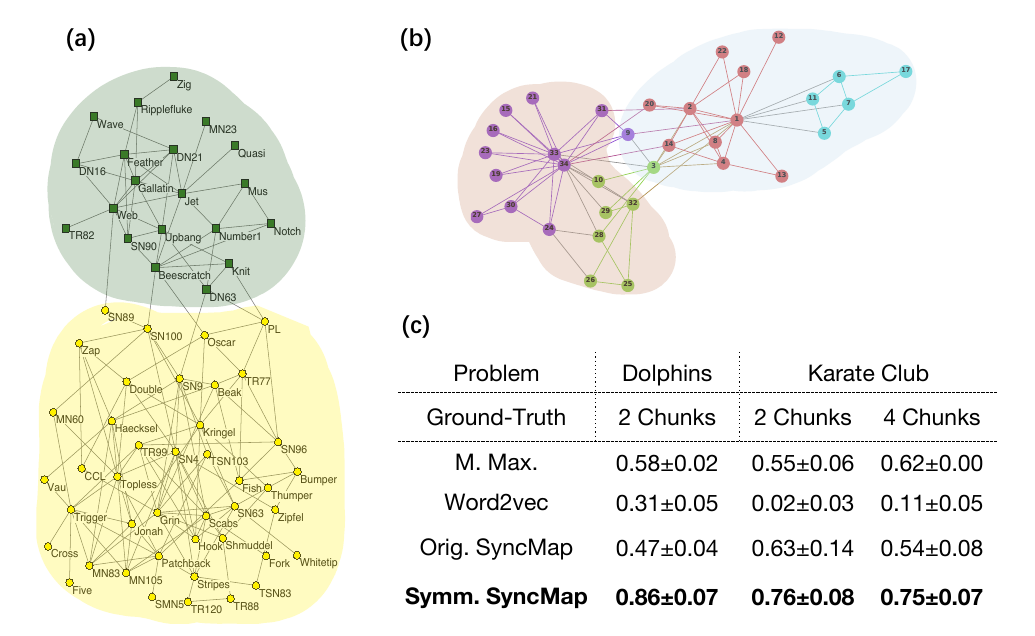}
    \caption{
    \textbf{Two community detection benchmarks used in experiments.} 
    (a) Dolphins network. Colors denote labels. Figure modified from Ref. \cite{arenas2008analysis}.  
    (b) Karate network. Colors of nodes denote local communities while colored shadow areas define the global communities. Figure modified from Ref. \cite{perozzi2014deepwalk}.
    (c) NMI results. 
    See  \ref{Appendix:realworld_analysis} for more analysis.
    }
  \label{fig:karate1}
\end{figure*} 

\begin{figure*}[ht]
    \centering
    \includegraphics[width=0.99\linewidth]{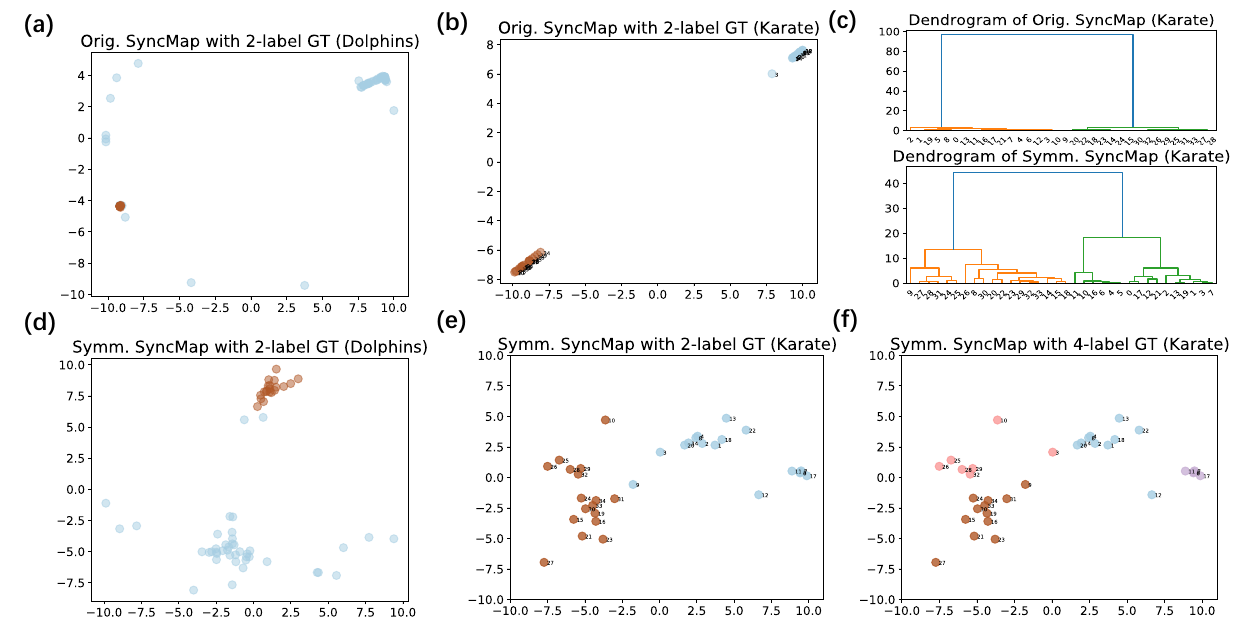}
    \caption{
    \textbf{Results of real-world scenarios with the original and Symmetrical SyncMap.} 
    (a) and (b) Original SyncMap's learned representations of Dolphins network and Karate network, respectively.
    (c) Dendrograms by hierarchical clustering (ward linkage) in Karate problems by Original SyncMap (top panel) and Symmetrical SyncMap (bottom panel).
    (d) and (e) Symmetrical SyncMap's learned representations of Dolphins network and Karate network, respectively.
    (f) Same as (e), while using 4 ground truth labeling.
    Colors indicate the true labels of the communities.
    See  \ref{Appendix:realworld_analysis} for more analysis.
    }
  \label{fig:karate2}
\end{figure*} 

\section{Real-world Scenarios}
\label{sec:real world}
\textcolor{black}{
We study four real-world scenarios to verify the performance of Symmetrical SyncMap: (1) a first-order Markov model of theme transitions for humpback whales’ song \cite{garland2017song} that was used in \cite{vargas2021syncmap}; (2) a network of stochastic block model (SBM); and (3) two social network datasets with well-established community structures. The ground-truth of (1) and (2) are defined in  \ref{Appendix:realworld_analysis}. Parameter settings of all models remained the same as in previous imbalanced CGCP experiments.
}

\textcolor{black}{
For the first problem, state transition in the humpback whales’ song are considered equally probable, as the transition between nodes are not given. As shown in Table \ref{table2}, the NMI of all models are relatively low, because of the difficulty to detecting overlapping states which make chunks not as distinct as those in other tests.
}
For the SBM, we test a reference network introduced by \cite{lee2019review}, where the network is considered as a graph which was then converted to a high-dimensional CGCP (i.e., 3 sightly imbalanced communities with a total of 90 nodes and 1192 edges). 
Both original and Symmetrical SyncMap yield the optimal solution (see Table \ref{table2}), showing the capabilities to tackle with large-scale CGCP.

We then test two well-studied benchmark networks in community detection. Hierarchical clustering was applied to replace DBSCAN in the clustering phase, to produce dendrograms for the visualization of hierarchies (i.e., by specifying the number of communities/chunks). Detailed settings and analysis for baselines are in  \ref{Appendix:realworld_analysis}. NMI results are shown in Figure \ref{fig:karate1}(c) with statistics. 

The first one is the Lusseau’s network of bottlenose dolphins \cite{fortunato2010community}, an imbalanced structure with 2 ground-truth communities of sizes 20 and 42. Our algorithm yields a much higher NMI than other algorithms. It avoids forming dense communities produced by the original SyncMap (see Figure \ref{fig:karate2}(a)(d)), allowing local relationships to be extracted, as verified in the following problem. 

The second problem is the Zachary’s karate club network which contains 34
nodes and 78 undirected and unweighted edges. 
We used two sets of ground-truth: (i) 2 chunks labeled by the original paper \cite{zachary1977information}, and (ii) 4 chunks found through modularity-based clustering \cite{perozzi2014deepwalk}.

Symmetrical SyncMap depicts the global graph structure while preserving the topology of local communities (Figure \ref{fig:karate2}(c)(e)(f)). In contrast, the original SyncMap can only separate two global communities with a very dense representation (Figure \ref{fig:karate2}(b)(e)). This unavoidable convergence is due to the stronger negative feedbacks over time, pulling away nodes from each community/chunk. 
Note that the representation learned by our method is comparable to node embeddings models with loss functions required and with more expensive training procedures such as DeepWalk \cite{perozzi2014deepwalk} and Graph Convolutional Networks \cite{kipf2016semi}. 
Unlike graph-based models, we achieve this by (i) mapping correlations from temporal input to a latent state space, (ii) keeping equilibrium by symmetrical activation (otherwise nodes would be locked in dense communities), and thus (iii) enabling hierarchies to be extracted from sequences.
More importantly, the inherent adaptivity, as shown in previous experiments, suggests that our model a has potential usage in \textit{inductive} applications, as it does not require any additional optimizations when dealing with new nodes/(sub)structures, while \textit{transductive} methods such as DeepWalk cannot naturally generalize to unseen nodes or changed structures \cite{hamilton2017inductive}.

Having said that, we argue that these real-world scenarios usually (i) have no ground-truth and (ii) are strongly biased towards standard algorithms. To illustrate, the absence of ground-truth has to do with the fact that: it is not only difficult to define the social structures, but also hard to know the existence of real chunks in nature; thus, any answer would be a guess at most. Besides, the bias is due to the output, used as ground-truth, is found by using standard algorithms in the original papers, which makes good results in real-world data more like “algorithms that perform similar to standard algorithms”, rather than “algorithms that work with real-world data”.



\section{Conclusions}
We propose Symmetrical SyncMap, a brain inspired self-organizing algorithm built on top of the original work to solve continual general chunking problems (CGCP).
Experiments of different CGCPs have illustrated how effective the concise modifications work on those challenging tasks. 
By applying symmetrical activation to the dynamical equations in which loss/optimization functions are not required, 
our algorithm not only learns imbalanced CGCP data structures with great long-term stability and adaptivity, but also shows the potentials to uncover complex hierarchical topologies encoded in temporal sequences.
This reveals the self-organizing ability of the proposed algorithm in analyzing the community structures of a broad class of temporal inputs.
Future goals, advised by the results presented in this paper, will be to investigate CGCP problems with large scale, hierarchies and noisy environments, and tasks specific to representation learning in various real-world scenarios.

\section{Acknowledgments}
This work was supported by JSPS Grant-in-Aid for Challenging Exploratory Research—Grant Number JP22534665, JST Strategic Basic Research Promotion Program (AIP Accelerated Research)—Grant Number 22584686, JSPS Research on Academic Transformation Areas (A)—Grant Number 22572551, JST SPRING—Grant Number JPMJSP2136.

\appendix

\section{Definition of Normalized Mutual Information (NMI)}
\label{Appendix:nmi}
Among the large number of clustering quality measurements, we used the  Normalized Mutual Information (NMI) for measurement. Mathematically, NMI is defined as:
\begin{equation} \label{eq8}
    { 
    NMI(\hat{Y, Y})=
    \frac{I(\hat{Y};Y)}{\frac{1}{2}(H(\hat{Y})+H(Y))}, \quad NMI \in [0,1]}
\end{equation}
\noindent where $\hat{Y}$ and $Y$ are the output of algorithms and the truth labels, respectively. $I(\hat{Y} ; Y )$ is the mutual information and $H(*)$ is the entropy. 
NMI ranges between 0 and 1, and the higher the score, the better the clustering performance 
(better correlation between $\hat{Y}$ and $Y$). 

\textcolor{black}{Regarding the choice of measurement, here are the main reasons of using NMI:
}
\begin{itemize}
    \item \textcolor{black}{Robustness to Label Permutation: NMI is inherently robust to label permutation. This means that even if the clusters are labeled differently in the true and predicted clusters, NMI will still provide an accurate measure of similarity. This property makes NMI especially suitable in our context, where the absolute labels of the clusters are not as important as the relative groupings.
    }
    \item \textcolor{black}{Handling of Unequal/Imbalanced Cluster Sizes: Error rates such as false positives and negatives can be heavily influenced by the size of the clusters. If some clusters are significantly larger than others, which is exactly our focus of the problem, they can dominate the error rates, leading to skewed results. NMI, on the other hand, is less affected by such size discrepancies, providing a more balanced and representative measure of performance.
    }
    \item \textcolor{black}{Applicability to Overlapping Clusters: In the CGCP settings, the nodes can belong to overlapping clusters, which cannot be effectively evaluated using false positive and false negative rates. NMI, however, can handle such scenarios, making it a more appropriate choice in our case.
    }

\end{itemize}

\section{Examples of Imbalanced CGCPs}
\begin{figure*}[ht]
    \centering
    \includegraphics[width=0.99\textwidth]{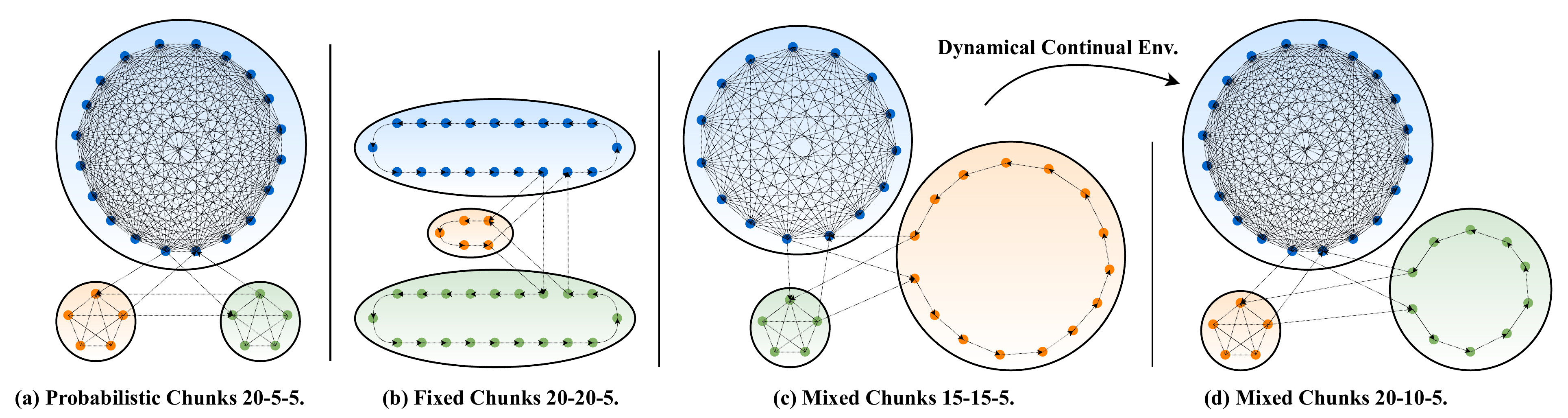}
    \caption{Examples of imbalanced chunking problems used in the experiments. 
    }
    \label{fig:chunk_problems_all}
\end{figure*} 

\label{Appendix:exmaple CGCP}
We show the examples of imbalanced chunking problems used in the experiments. 
In Figure \ref{fig:chunk_problems_all}(a), probabilistic chunks are shown, which is a graph structure that allows random walking. 
Figure \ref{fig:chunk_problems_all}(b) are fixed chunks: temporal chunks defined originated from neuroscience.
Figure \ref{fig:chunk_problems_all}(c)(d) are mixed chunks: integration of fixed and probabilistic chunks. 
In the continual setting, the causal structure can change over time. 
Dots inside each circle belong to a corresponding chunk. 
Lines connecting nodes without an arrow indicate that the transition is bidirectional; for directional transitions, arrows specify the direction.

\section{Analysis of Real-world Scenarios}
\label{Appendix:realworld_analysis}

\begin{figure*}[ht]
    \centering
    \includegraphics[width=0.95\textwidth]{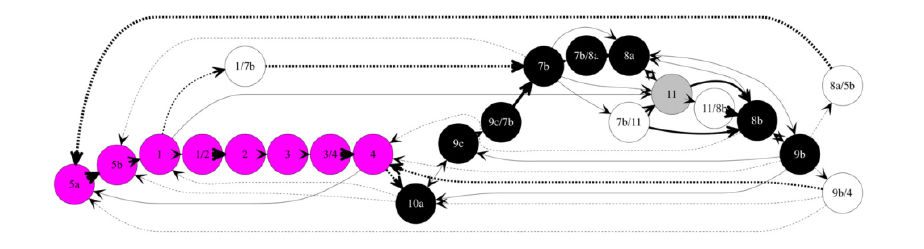}
    \caption{\textcolor{black}{First-order Markov Chain of humpback whales’ song types \cite{garland2017song}. Since the transition between nodes are not given, they are considered equally probable (uniform).
    }
    }
    \label{fig:seq1}
\end{figure*} 
\begin{figure*}[ht]
    \centering
    \includegraphics[width=0.99\textwidth]{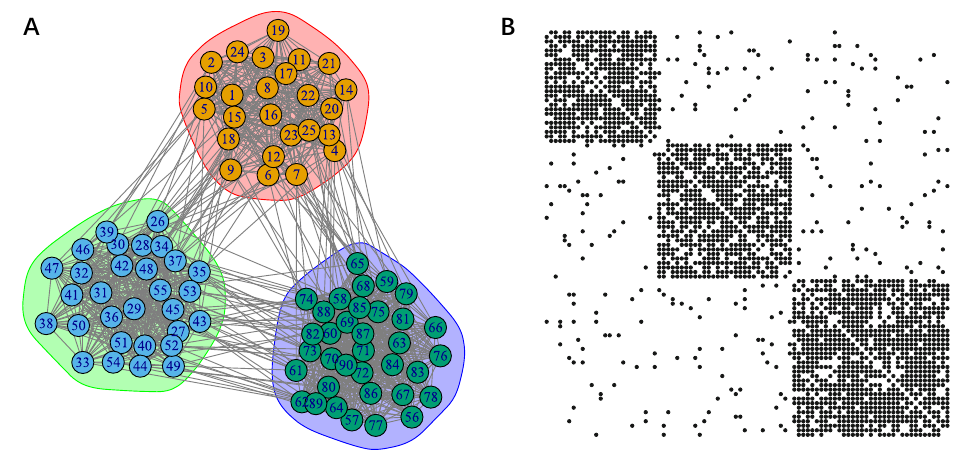}
    \caption{Stochastic block model (SBM) network used in experiments. (a) Graph introduced in \cite{lee2019review}, which consists of 90 nodes and 1192 edges. The nodes are divided into 3 groups, with groups 1, 2 and 3 containing 25, 30 and 35 nodes, respectively. The nodes within the same group are more closely connected to each other, than with nodes in another group. The connectivity of the nodes is considered uniform transition. (b) Corresponding adjacency matrix for graph in (a).
    }
    \label{fig:SBM}
\end{figure*} 
\begin{figure*}[ht]
    \centering
    \includegraphics[width=0.9\textwidth]{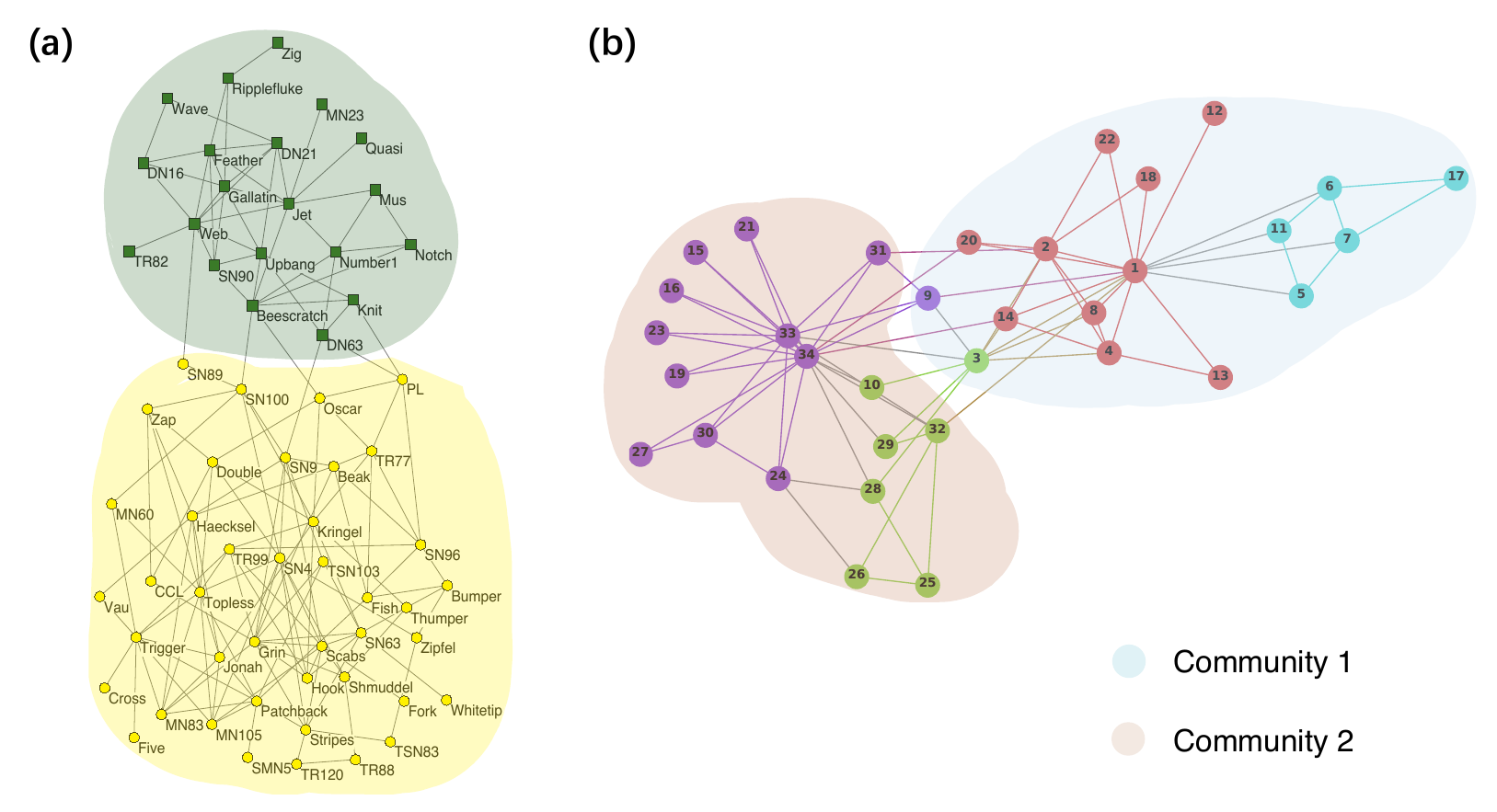}
    \caption{Two community detection benchmarks used in experiments. 
    (a) The Lusseau’s network of bottlenose dolphins \cite{fortunato2010community}, modified from \cite{arenas2008analysis}. This network is considered as an imbalanced structure with 2 ground-truth communities of sizes 20 and 42.
    (b) The  Zachary’s karate club network, modified from \cite{perozzi2014deepwalk}. This contains 34 nodes and 78 undirected and unweighted edges. Colors within nodes denote local communities (i.e., the ground-truth found by the modularity-based algorithm \cite{perozzi2014deepwalk}), while colored shadow areas define the global communities (i.e., the ground-truth collected from the original paper \cite{zachary1977information}).
    }
    \label{fig:kc_dol_graph}
\end{figure*} 

\subsection{Experiment settings of models used in real-world scenarios}
Here, we specify the detailed settings in the experiments of real-world problems.

\textcolor{black}{\textbf{Humpback whales’ song.} The structure of humpback whales’ song is shown in Figure \ref{fig:seq1}.
As mentioned in the main text, all settings remained the same as in previous imbalanced CGCP experiments (see Section \ref{sec:experiment}).
}

\textbf{Stochastic block model (SBM) network.} The structure of SBM network is shown in Figure \ref{fig:SBM}. All settings remained the same as in previous imbalanced CGCP experiments (see Section \ref{sec:experiment}).

\textbf{Community detection benchmarks.} 
In the two problems of the community detection benchmarks, MRIL was not used, since we focus on the investigation of how the given algorithms learn the topological and hierarchical structures underlying in input sequences, as well as how well the structures are produced using such algorithms. MRIL cannnot encode input sequences into a map space, therefore it was not considered as a baseline.

Regarding the Modularity Max, we again used TP matrices that produced by the graphs generated from the given input sequences, as our focus is on the ability of extracting information in input sequences.
When finding the communities, we specify the ``number of communities" to the algorithm. This can be found in the official documentation of NetworkX (a Python library for analyzing graphs), where we set ``a minimum number of communities below which the merging process stops. The process stops at this number of communities even if modularity is not maximized."
However, it should be noted that the process will stop before the cutoff if it finds a maximum of modularity. 
Based on the given ground-truth, in the Lusseau’s network of bottlenose dolphins, we set the ``number of communities" at 2. In the Zachary’s karate club network, we set the ``number of communities" at 2 and 4 for computing NMI with two sets of ground-truth, respectively.

Regarding the Word2vec, we set the latent dimension equal to 2, and kept all other parameter settings unchanged. This is to produce a 2-D representation of the given community structures. Also, in the previous experiments, DBSCAN was used to obtain chunks, where in the real-world problems we replaced it to hierarchical clustering as we are more interested in the topology as well as the hierarchical structures.

The original SyncMap and Symmetrical SyncMap shared the same changes with Word2vec, that is, we only reduced the SyncMap space dimension $k$ from 3 to 2. Also, hierarchical clustering is used.

Regarding the hierarchical clustering, we used ``ward" as a linkage method. And we specified the number of clusters when performing the algorithm. In details, for the Lusseau’s network of bottlenose dolphins, we set the ``number of clusters" at 2. In the Zachary’s karate club network, we set the ``number of clusters" at 2 and 4 for computing NMI with two sets of ground-truth, respectively.

\subsection{Results analysis}

\textbf{SBM network.} 
Modularity Max performs nearly optimal in this scenario. Also, recall that it yields relatively low NMI in fixed CGCP. The differences of the performance observed here might be because the communities with many deterministic connections are less likely to appear in usual problems faced by Modularity Max, thus leading to a problem bias; that is, fixed chunk structures are not strictly meet the condition that the possibility of transition to an internal state within a chunk is higher than that of transition to an external state belonging to other chunks. This leads to a problem bias for Modularity Max.

\begin{figure*}[ht]
    \centering
    \includegraphics[width=0.9\textwidth]{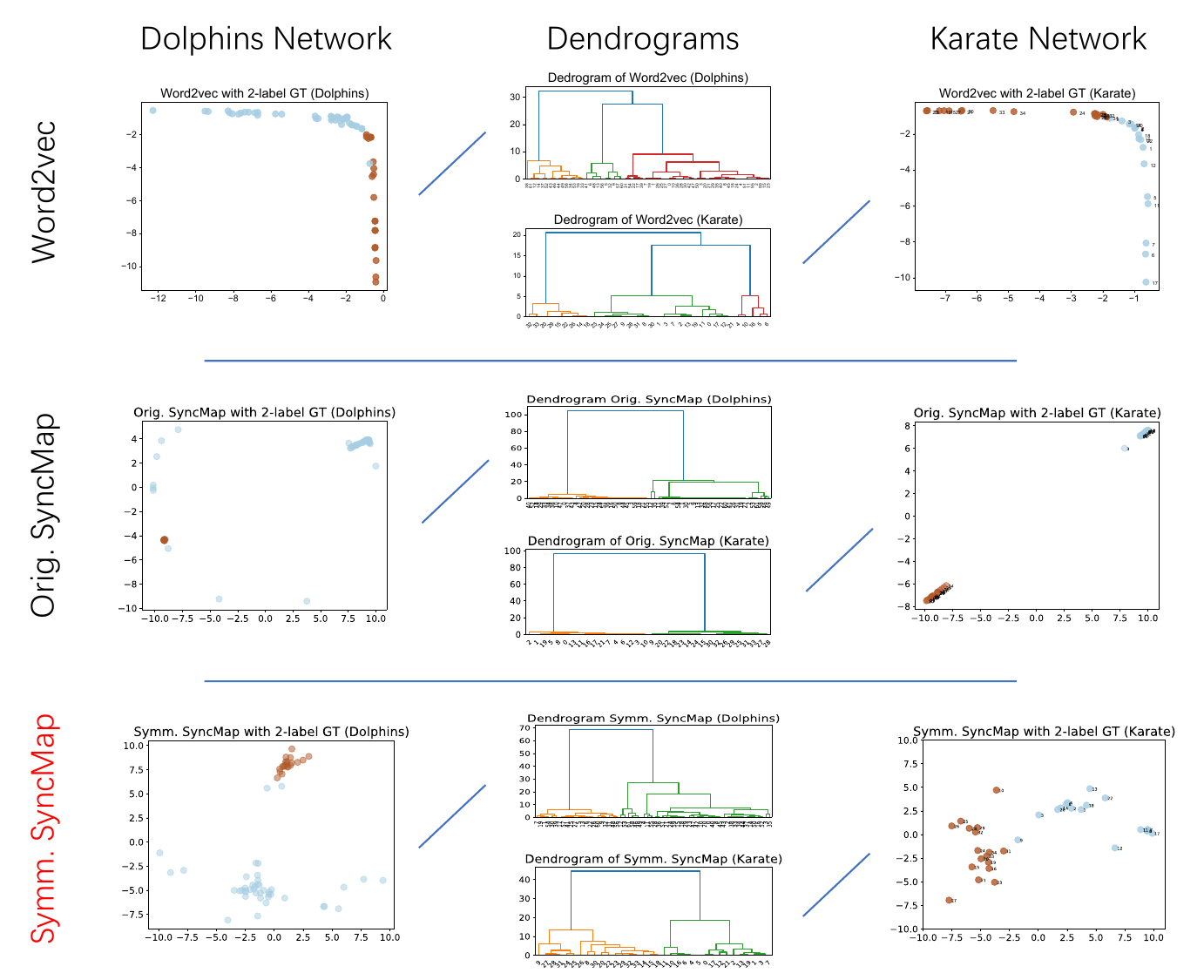}
    \caption{Analysis of real-world problems used in the experiments. Here we compare three algorithms, namely (i) Word2vec, (ii) original SyncMap, and (iii) Symmetrical SyncMap (the proposed one). Note that Modularity Max and MRIL do not encode input sequences into a map space. Thus, they are not figuratively comparable.
    }
    \label{fig:realworldappendix}
\end{figure*} 

\textbf{Community detection benchmarks.}
Word2vec, as shown in Figure \ref{fig:realworldappendix}, always forms a latent representation with two long tails. This shape does harm when finding hierarchical structures.

We have analyzed the two models of SyncMap in the main text. It is worth noting here that from the dendrograms of the original SyncMap, we can conclude that the original work is not possible to extract local relationships (i.e., hierarchies), as the nodes are all densely distributed in compact communities.

Regarding the Modularity Max, it does not encode temporal sequences into a map space, and thus it is not figuratively visualized.
Meanwhile, although the datasets here are designed for modularity-based models, \textbf{Modularity Max} failed to extract information and maximize the modularity from sequential data, thus showing lower NMI.

\section{Maps learned by Different Algorithms}
\label{Appendix:learned map}
\begin{figure*}[ht]
    \centering
    \includegraphics[width=0.99\textwidth]{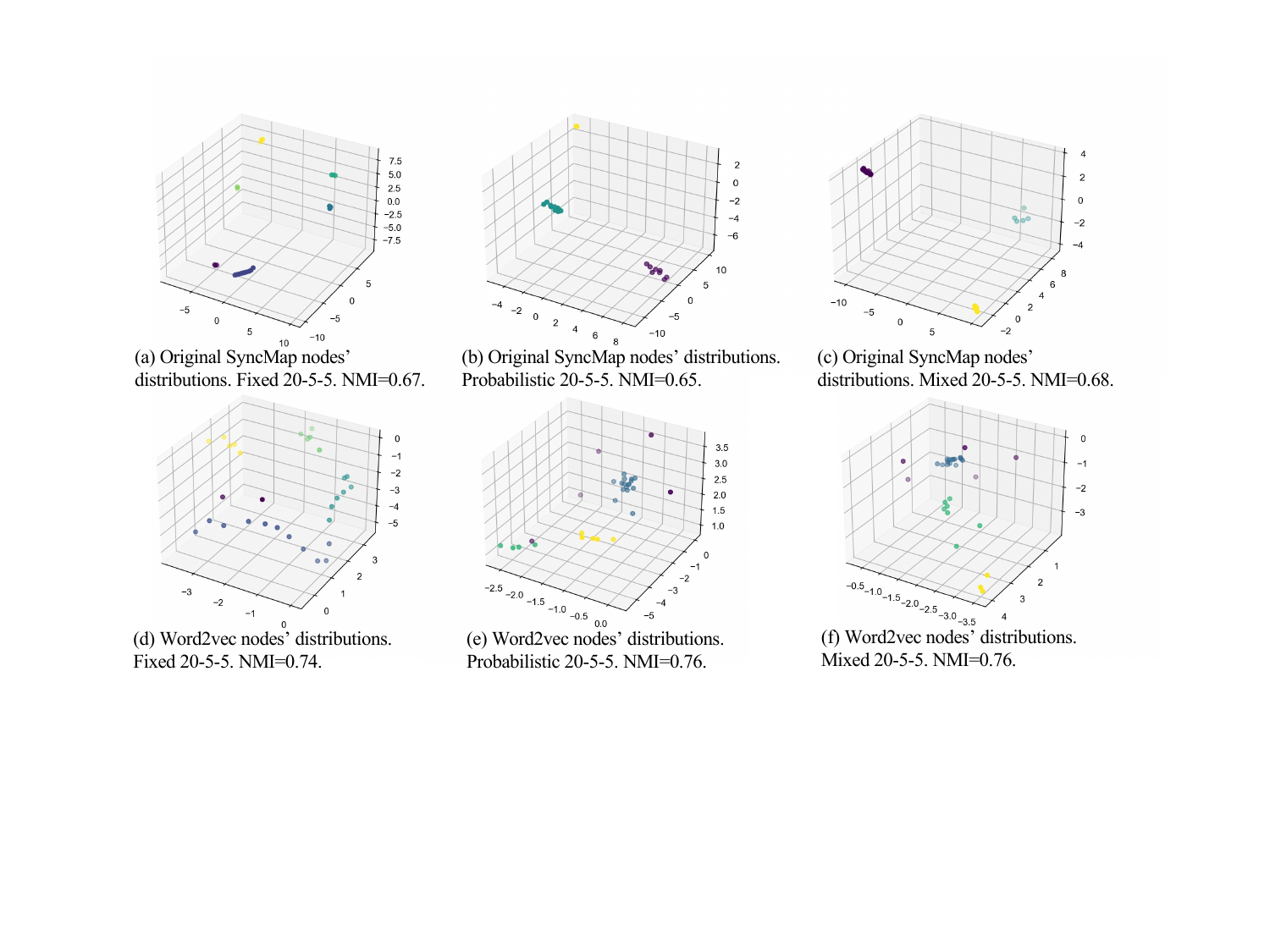}
    \caption{Maps learned by Different Algorithms. 
    }
    \label{fig:UB_figures}
\end{figure*}

Figure \ref{fig:UB_figures} shows the chunking results of imbalanced 20-10-5 fixed (left), probabilistic (middle) and mixed (right) problems using the original SyncMap (top), Word2vec (middle) and the proposed Symmetrical SyncMap (bottom). Colors of the nodes indicate the true labels of chunks. Specifically, purple for big chunk (20), yellow for moderate chunk (10), green for small chunk (5). 
Note that Modularity Max and MRIL do not encode input sequences into a map space. Thus, they are not figuratively comparable.

\newpage
\section{Stability Analysis}
\label{Appendix:stability_analysis}

\begin{figure*}[ht]
    \centering
    \includegraphics[width=0.83\textwidth]{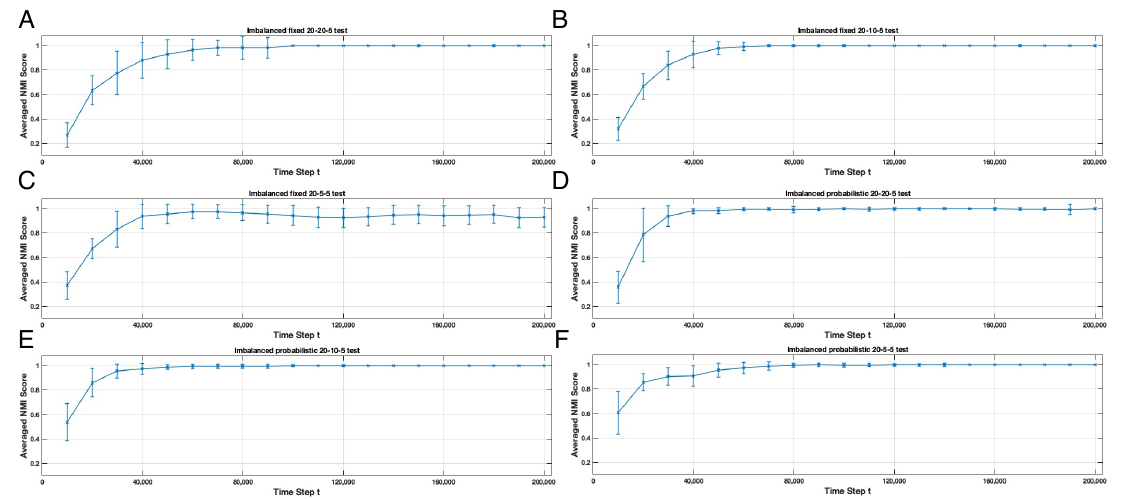}
    \caption{NMI over time of imbalanced tests. 
    (a) fixed 20-20-5, 
    (b) fixed 20-10-5, 
    (c) fixed 20-5-5, 
    (d) probabilistic 20-20-5, 
    (e) probabilistic 20-10-5, 
    (f) probabilistic 20-5-5. 
    }
    \label{fig:stability1}
\end{figure*}

\begin{figure*}[ht]
    \centering
    \includegraphics[width=0.83\textwidth]{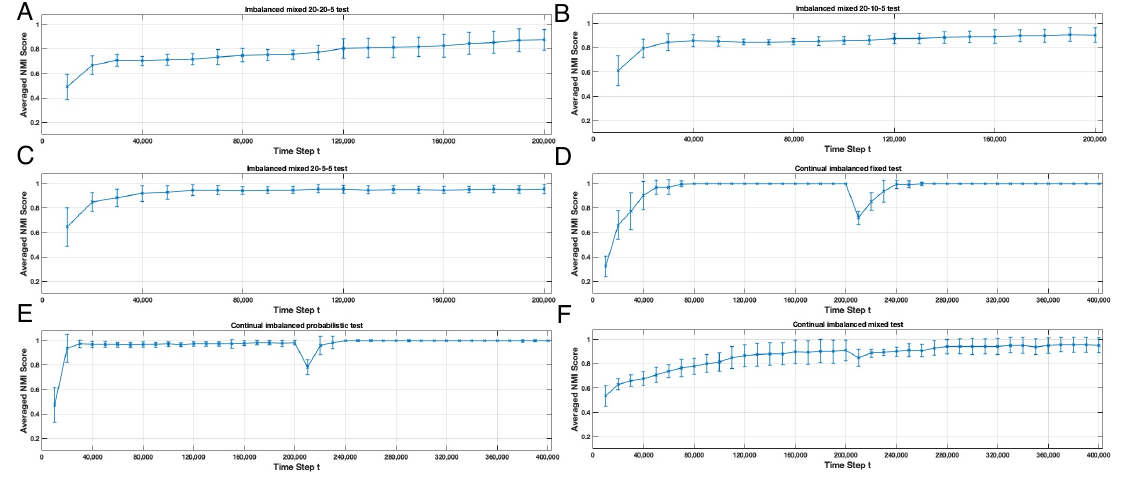}
    \caption{NMI over time of imbalanced tests. 
    (a) mixed 20-20-5, 
    (b) mixed 20-10-5, 
    (c) mixed 20-5-5, 
    (d) continual fixed, 
    (e) continual probabilistic, 
    (f) continual mixed. 
    }
    \label{fig:stability2}
\end{figure*}

The proposed Symmetrical SyncMap shows good stability over the long run in all imbalanced chunking problems. Figures \ref{fig:stability1} and \ref{fig:stability2} show the NMI average with error bar (s.t.d.) at every 10,000 time step during training in all imbalanced problems.


\newpage
\newpage

\section{Parameter Sensitivity Analysis}
\label{Appendix:Parameter Sensitivity Anaylsis}

\subsection{Choosing the threshold factor ``$a$''}
\begin{figure*}[ht]
    \centering
    \includegraphics[width=0.6\textwidth]{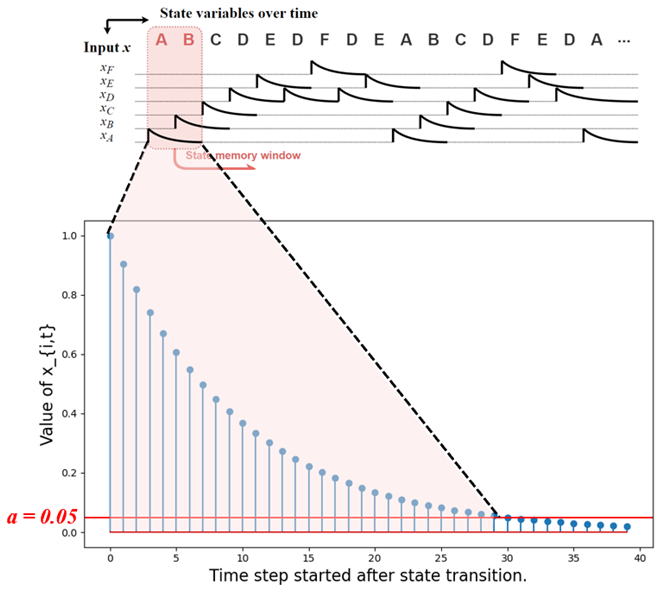}
    \caption{
    \textcolor{black}{Explanation of the thresholding process. 
    }
    }
    \label{fig:thresholding}
\end{figure*}

\textcolor{black}{Both Symmetrical SyncMap and the original works use (1) exponential-decaying encoding, and (2) time delay $tstep$ for generating the input vector at time step t, i.e., 
$\boldsymbol{x}_t={\{x_{1,t},...,x_{i,t},...,x_{N,t}\}}^{T}$. 
As shown in Figure \ref{fig:thresholding} in the following, that for a particular state variable, for example $x_{i,t}$, when it is activated (i.e., state transitioning into state $i$), the value of $x_{i,t}$ is exponentially decreasing according to Equation 6 in the manuscript. Consider the case of time delay $tstep=10$, and the state memory $m=3 $, we can easily compute the threshold factor $a=0.05$. In other words, in order to let the system activate state $i$ for $m\ast tstep=30$ steps, the threshold factor is set to $a=0.05$. 
}

\textcolor{black}{In fact, generalizing the memory helps to capture the fixed chunks. As shown in Appendix D (left-bottom panel), SyncMap forms a fixed chunk in its space in a curvy manner. Having a bigger memory window (i.e., decrease the value of $a$) can make the nodes belonging to the fixed chunk more compact.
}

\subsection{Parameter robustness analysis}
Table \ref{parameter_robustness} shows the performance of Symmetrical SyncMap on the same imbalanced experiments (fixed/probabilistic/mixed 20-10-5 test) but with different parameters settings. Results suggest that Symmetrical SyncMap is robust to changes in parameters, with mostly smooth changes.

Note that the parameter $Pr$ is designed for solving small chunks. To illustrate, if $Pr = 100\%$, then at every time step we apply state memory generalization to have a longer memory window. In this case, if the current chunk is too small (e.g., only having 3 states), then it would probably not be detected. Therefore, choosing $Pr=30\%$ provides a trade-off between finding big and small chunks. In fact, in the future version of our model, this parameter will no longer be used, as tiny chunk/communities are rarely appear alone, while they usually appear inside a big chunk (i.e., hierarchies). Having said that, the results of using $Pr=100\%$ and $Pr=30\%$
are both adequate.

\begin{table*}[ht]
    \centering
    \caption{NMI of several Symmetrical SyncMap variations in Fixed, Probabilistic and Mixed structures settings (20-10-5 tests).
    }
    \scalebox{0.9}{
    \begin{tabular}{c c c |c c c}
    \hline
    \multicolumn{3}{c|}{\textbf{Setting}} & {\textbf{Fixed}} & {\textbf{Probabilistic}} & {\textbf{Mixed}} \\
    \hline
      \textbf{k = 3} & \textbf{m = 3} &	\textbf{$\boldsymbol{Pr}$ = 30\%} & 1.0±0.0 & 1.0±0.0 & 0.92±0.06	\\
        
     \textbf{k = 3} & \textbf{m = 2} &	\textbf{$\boldsymbol{Pr}$ = 100\%} & 0.95±0.08 & 0.98±0.02 & 0.86±0.03\\
    \textbf{k = 3} & \textbf{m = 3} &	\textbf{$\boldsymbol{Pr}$ = 100\%} & 1.0±0.0 & 1.0±0.0 & 0.95±0.05\\
    \textbf{k = 3} & \textbf{m = 4} &	\textbf{$\boldsymbol{Pr}$ = 30\%} & 0.96±0.08 & 1.0±0.0 & 0.95±0.05\\
     \textbf{k = 2} & \textbf{m = 2} &	\textbf{$\boldsymbol{Pr}$ = 100\%} & 0.90±0.10 & 1.0±0.0 & 0.87±0.07\\
     \textbf{k = 2} & \textbf{m = 3} &	\textbf{$\boldsymbol{Pr}$ = 30\%} & 0.98±0.07 & 1.0±0.0 & 0.90±0.08\\
     \textbf{k = 2} & \textbf{m = 4} &	\textbf{$\boldsymbol{Pr}$ = 30\%} & 0.95±0.08 & 1.0±0.0 & 0.94±0.07\\
      \hline
    \end{tabular}}
    \label{parameter_robustness}
\end{table*}

\section{Statistical Tests}
\label{Appendix:Statistical Tests}
We used a t-test with p-value of 0.05 to verify if the best result is statistically significantly different from other results. $h$ is the hypothesis test result ($h$=0 indicates a failure to reject the null hypothesis at the 5\% significance level, and $h$=1 otherwise). $p$ is the two-tailed $p$ value, and $ci$ is the confidence interval for the difference in population means of two samples.

\begin{table*}[ht]
    \centering
    \caption{Statistical Results.}
    \scalebox{0.6}{
    \begin{tabular}{|c |c |c |c |c|}
    \hline
    \textbf{Problems} & \textbf{Description} & \makecell[c]{\textbf{Num. of Samples} \\ \textbf{and Mean+s.t.d.}} &
    \textbf{Test} & \textbf{Statistic}\\
    \hline
      \makecell[c]{Long-term Analysis \\ (Figure 5)} 
      & Orig. vs Symm.SyncMap 
      & \makecell[c]{30 each model \\ at final time step} 
      &	\makecell[c]{Two-sample \\ t-test}
      &	\makecell[c]{h=1, \textbf{p=6.5228e-27}, \\ ci=[0.3166;0.3899]} \\
    \hline
      \makecell[c]{Prob. 20-20-5 \\ (Table 1)} 
      & M.Max vs Symm.SyncMap
      & \makecell[c]{30 each model \\(0.96±0.04 and 1.0±0.0)} 
      &	\makecell[c]{Two-sample \\ t-test}
      &	\makecell[c]{h=1, \textbf{p=4.3546e-06}, \\ ci=[0.0220;0.0508]} \\      
    \hline
      \makecell[c]{Prob. 20-20-5 \\ (Table 1)} 
      & Orig. vs Symm.SyncMap
      & \makecell[c]{30 each model \\(1.0±0.0 and 1.0±0.0)} 
      &	\makecell[c]{Two-sample \\ t-test}
      &	\makecell[c]{h=NaN, \textbf{p=NaN}, \\ ci=[0.0;0.0]} \\  
    \hline
      \makecell[c]{Prob. 20-10-5 \\ (Table 1)} 
      & M.Max vs Symm.SyncMap
      & \makecell[c]{30 each model \\(1.0±0.0 and 1.0±0.0)} 
      &	\makecell[c]{Two-sample \\ t-test}
      &	\makecell[c]{h=0, \textbf{p=0.3215}, \\ ci=[-0.0033;0.0098]} \\  
    \hline
      \makecell[c]{Prob. 20-5-5 \\ (Table 1)} 
      & M.Max vs Symm.SyncMap
      & \makecell[c]{30 each model \\(1.0±0.0 and 1.0±0.0)} 
      &	\makecell[c]{Two-sample \\ t-test}
      &	\makecell[c]{h=0, \textbf{p=0.3216}, \\ ci=[-0.0042;0.0125]} \\  
    \hline
      \makecell[c]{Mixed 20-20-5 \\ (Table 1)} 
      & Orig. vs Symm.SyncMap
      & \makecell[c]{30 each model \\(0.84±0.08 and 0.87±0.09)} 
      &	\makecell[c]{Two-sample \\ t-test}
      &	\makecell[c]{h=0, \textbf{p=0.1568}, \\ ci=[-0.0121;0.0731]} \\  
    \hline    
      \makecell[c]{SBM Network \\ (Table 1)} 
      & M.Max vs Symm.SyncMap
      & \makecell[c]{30 each model \\(0.99±0.02 and 1.0±0.0)} 
      &	\makecell[c]{Two-sample \\ t-test}
      &	\makecell[c]{h=0, \textbf{p=0.09}, \\ ci=[-0.0011;0.0129]} \\        
    \hline
      \makecell[c]{SBM Network \\ (Table 1)} 
      & Orig. vs Symm.SyncMap
      & \makecell[c]{30 each model \\(1.0±0.0 and 1.0±0.0)} 
      &	\makecell[c]{Two-sample \\ t-test}
      &	\makecell[c]{h=NaN, \textbf{p=NaN}, \\ ci=[0.0;0.0]} \\  
    \hline
    \end{tabular}}
    \label{table:stat}
\end{table*}

\section{Computational Time Analysis}
\label{Appendix:Computational Time Anaylsis}
We analyze the computational time over Symmetrical SyncMap, SyncMap and Word2vec. In addition to Karate network, we introduced two larger scale imbalanced CGCP problems (i.e., problem with 300-D input and 1200-D input). More specifically, the structure of 300-D CGCP problem includes: Four chunks with 50 states + Two chunks with 20 states + Four chunks with 10 states + Four chunks with 5 states. We denote this problem as 300-D (50x4 + 20x2 + 10x4 + 5x4).
Using the similar denotation, the 1200-D CGCP problem is 1200-D (300x2 + 150x2 + 50x4 + 20x2 + 10x4 + 5x4).

We ran all three problems 10-time per problem per algorithm with 200,000 as sequence length. All tests were run on a MacBook Pro 2.4GHz Quad-Core Intel Core i5 16GB laptop as they demand little computational effort, and computation time [second] were obtained in mean±s.t.d.
Results show that the proposed Symmetrical SyncMap is scalable. Although it is slower than the original one and Word2vec (mainly due to the stochastic selection process), the computation time does not become worse as the scale of the input increases. Note that both SyncMap and the proposed algorithm should improve considerably if parallelization, GPU programming and other techniques are employed. For example, all nodes in SyncMap can be updated at the same time.

\begin{table*}[ht]
    \centering
    \caption{Computation time[s] comparison over Symmetrical SyncMap, SyncMap and Word2vec.
    }
    \scalebox{0.99}{
    \begin{tabular}{ccccl}
    \cline{1-4}
    Problem Type      & Symm. SyncMap & Orig. SyncMap & Word2vec &  \\ \cline{1-4}
    Karate (34-D)     & 90.227±1.297                   & 30.179±0.571       & 33.077±0.429        &  \\
    Imbalanced 300-D  & 98.055±3.689                   & 34.462±0.671       & 43.035±0.494        &  \\
    Imbalanced 1200-D & 118.602±1.629                   & 44.827±0.979       & 72.956±1.923        &  \\ \cline{1-4}
    \end{tabular}
    }
    \label{computation_time}
\end{table*}


\section{Ablation Study} \label{Appendix:Ablation Study}
The proposed method is composed of two parts compared to the original SyncMap: Symmetrical activation and genalized memory window. We perform an ablation study to evaluate the effect of each of these two modifications.
The ablation study investigated CGCP problem of mixed-20-10-5. The key feature of this problem is that there is a fixed chunk with 10 state in between two probabilistic chunks. The 10-time averaged NMI results are shown in Table \ref{tab:ablation}.

\begin{table*}[ht]
    \centering
    \caption{NMI score of ablation study.
    }
    \scalebox{0.99}{
    \begin{tabular}{cc}
    \hline
    Model Type                                   & NMI Score \\ \hline
    Original SyncMap (m=2)                       & 0.84±0.07         \\
    SyncMap with Generalized Memory Window (m=3) & 0.83±0.0         \\
    SyncMap with Symmetrical Activation (m=2)    & 0.84±0.03         \\
    Symmetrical SyncMap (m=3)                    & 0.90±0.06         \\ \hline
    \end{tabular}
    }
    \label{tab:ablation}
\end{table*}

\begin{figure*}[ht]
    \centering
    \includegraphics[width=0.99\textwidth]{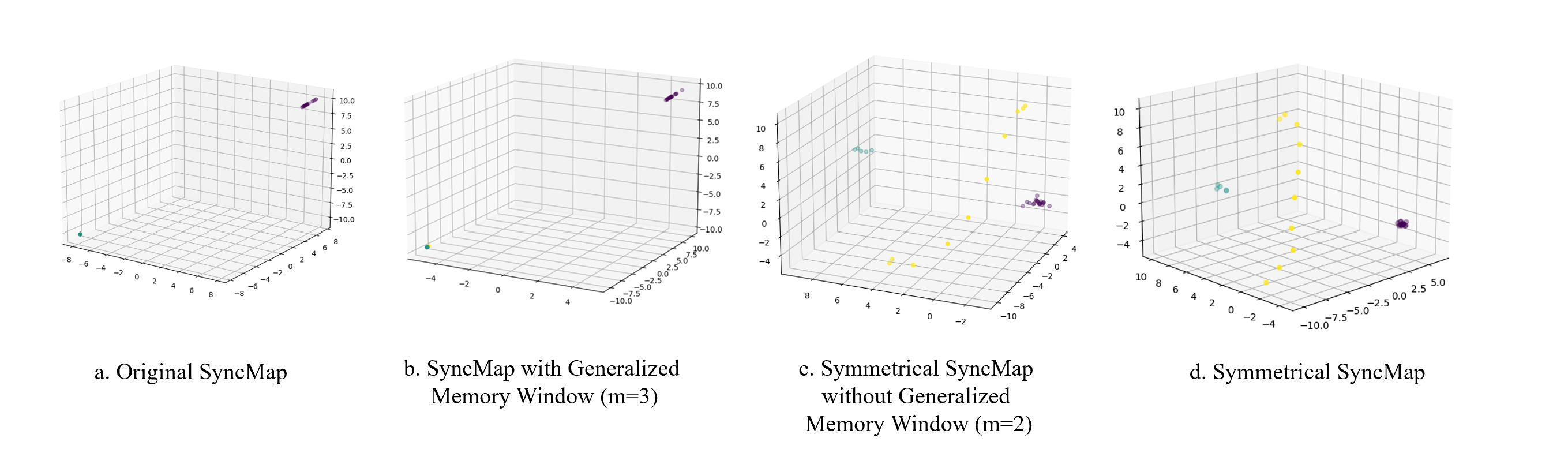}
    \caption{Learned map of (a) Original SyncMap, (b) SyncMap with Generalized Memory Window (m=3), (c) Symmetrical SyncMap without Generalized Memory Window (m=2), and (d) Symmetrical SyncMap. Note the nodes belong to fixed chunk is denoted by yellow color, which is slightly hard to visualize.
    }
    \label{fig:ablation}
\end{figure*}

From the NMI scores, it is hard to image what happen inside the SyncMap space. However, a better visualization can be observed in the learned map comparison in Figure \ref{fig:ablation}. 
To illustrate briefly, if only applying symmetrical activation, the states (i.e., weight nodes in the map) of the given fixed chunk will be very sparely distributed (Figure \ref{fig:ablation}(c), i.e., the system cannot detect the fixed chunk because it needs more memory to remember fixed chunks). On the other hand, if only applying generalized longer memory window, the strength of the negative feedback will still dominate the self-organizing process, thus resulting in the similar output with Original SyncMap (Figure \ref{fig:ablation}(a)(b); that is, all weight nodes are squeezed into compact clusters. Therefore, by applying symmetrical activation and generalized memory window together, the system is stable and able to detect fixed chunk in more general cases (Figure \ref{fig:ablation}(d).

\section{Related Works}
\label{Appendix:Related Works}
Chunking is an extremely multidisciplinary problem. In the main text, we attempt to cover a number of related topics from neuroscience to machine learning and try to connect them to this work.
Due to the page limit, we review additional related works here for completeness.

\textbf{Latent Variable Estimation.} Some literature focusing on latent variable estimation solve problems which are related to chunking \cite{fox2011sticky,qian2014learning,pfau2010probabilistic}. 
However, they have different objectives, since chunking is a self-organizing process over the variables of the problem with respect to their temporal correlation.
Even if chunks of variables can be abstracted as a set of variables, there is still an inherent difference between chunks and latent variables.


\section{Algorithmic Description of Symmetrical SyncMap}
\label{Appendix:Algorithmic}
\begin{center}
\scalebox{0.99}{
\begin{minipage}{1\linewidth}
\begin{algorithm}[H]
\caption{Symmetrical SyncMap} \label{alg1}
\textbf{Input}: input  sequence $\boldsymbol{X}=\{\boldsymbol{x_i}|i=1,...,\tau\}$ \\
\textbf{Parameters}: sequence length $\tau$, map dimension $k$, state memory $m$, probability parameter $Pr$, input dimension $n$, learning rate $\alpha$\\
\textbf{Output}: A number of clusters indicating communities and chunks.
\begin{algorithmic}[1]
\STATE Initialize SyncMap by generating weight nodes $w_{i,0}$ , where $i=1...n$
\STATE Set $W_t=\{w_{i,t}|i=1...n,t=0...\tau\}$ 
\FOR{$t=0$ to $\tau$}
\STATE Initialize $PS_t$ and $NS_t$ as empty set
\STATE Randomly generate a constant variable $Pr_t\in [0,1]$
\STATE Divide $m$ nodes into temporary set $PS_{temp}$
\STATE $m_{neg} \gets m$ (Symmetrical activation)
\IF{$m>2$ \& $Pr_t<Pr$}
    \STATE Stochastically select 2 nodes in $PS_{temp}$ to activate
    \STATE Include these 2 nodes into set $PS_{t}$
    \STATE $m_{neg} \gets 2$ (Symmetrical activation)
\ELSIF{$m>2$ \& $Pr_t\geq Pr$}
    \STATE $PS_{t} \gets PS_{temp}$
\ENDIF
\STATE Set temporary $NS_{temp} \gets W_t-PS_t$
\STATE Stochastically select $m_{neg}$ nodes in $NS_{temp}$ to activate
\STATE Include these $m_{neg}$ nodes into set $NS_{t}$
\STATE $PS_t$ and $NS_t$ determined
\STATE Calculate $cp_t$ and $cn_t$ by Eq. 2
\STATE Update the nodes' position by Eqs. 3 and 4 
\STATE Normalize nodes in hypersphere radius $=$ 10
\ENDFOR
\STATE Apply clustering algorithm such as DBSCAN or Hierarchical clustering
\end{algorithmic}
\end{algorithm}
\end{minipage}
}
\end{center}

\newpage
 \bibliographystyle{elsarticle-num} 
 \bibliography{arxiv.bbl}





\end{document}